\documentclass{article}
\usepackage[table,xcdraw]{xcolor}
\usepackage[preprint]{corl_2026}

\makeatletter
\if@conferencefinal
\renewcommand{\@noticestring}{%
  $^{\dagger}$Corresponding author. $^{*}$Equal advising.\par
  \@conferenceordinal\/ Conference on Robot Learning (CoRL \@conferenceyear), \@conferencelocation.%
}
\else
\if@preprinttype
\renewcommand{\@noticestring}{%
  $^{\dagger}$Corresponding author. $^{*}$Equal advising.%
}
\fi
\fi
\makeatother

\usepackage{xspace}
\usepackage{amsmath,amssymb,booktabs,tabularx,enumitem,wrapfig,graphicx}
\definecolor{y2rblue}{HTML}{4A7BA6}
\definecolor{y2ryellow}{HTML}{B8923A}
\definecolor{y2rsage}{HTML}{7FB069}
\definecolor{y2rcoral}{HTML}{C97B7B}
\definecolor{y2rmustard}{HTML}{E8B84A}
\definecolor{y2rlavender}{HTML}{9B7BB8}

\newcommand{\pname}{\texttt{LUCID}\xspace}

\title{\pname: Learning Embodiment-Agnostic Intent Models from Unstructured Human Videos for Scalable Dexterous Robot Skill Acquisition}

\author{
  Harsh Gupta$^{\dagger}$\\
  University of Illinois Urbana-Champaign\\
  \texttt{hgupt3@illinois.edu} \\
  \And
  Guanya Shi$^{*}$\\
  Carnegie Mellon University\\
  \texttt{guanyas@andrew.cmu.edu} \\
  \And
  Wenzhen Yuan$^{*}$\\
  University of Illinois Urbana-Champaign\\
  \texttt{yuanwz@illinois.edu} \\
}

\begin{document}
\maketitle

\begin{figure}[h]
  \vspace{-1em}
  \centering
  \includegraphics[width=\linewidth]{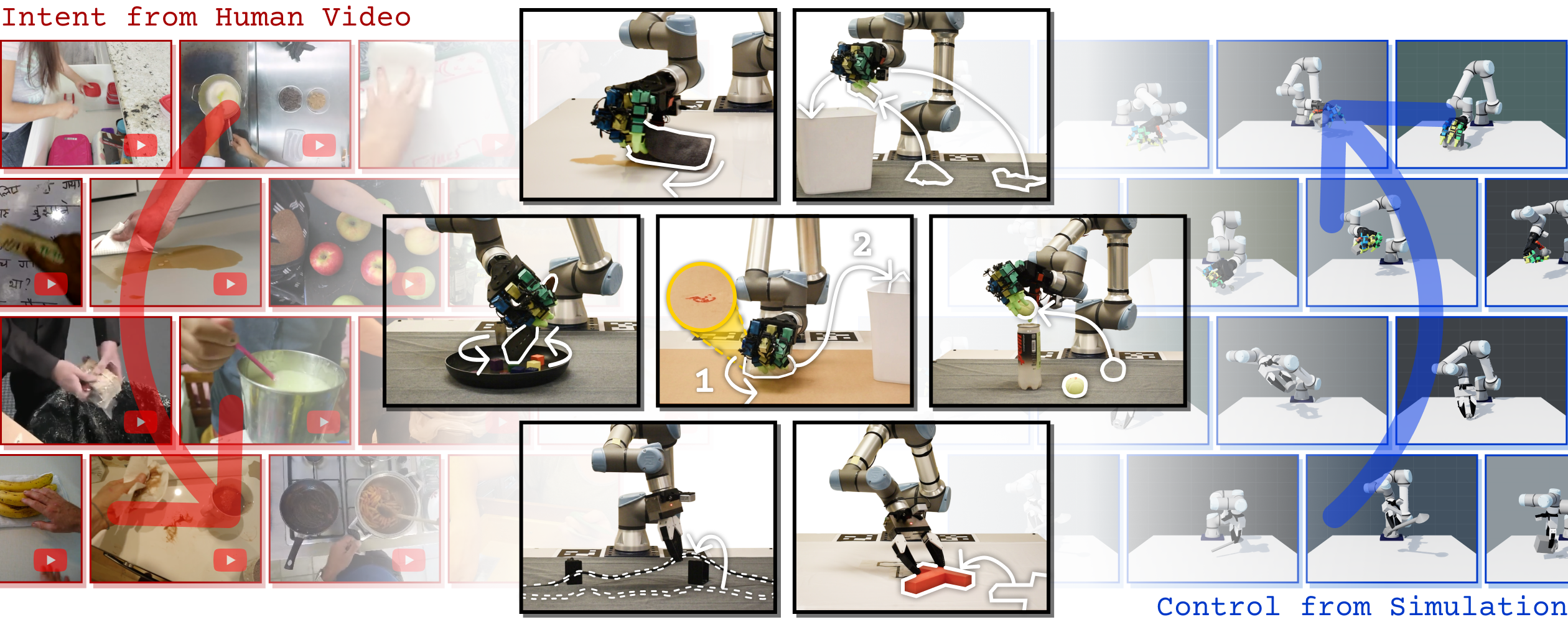}
  \caption{\textbf{\pname{}.} We learn a manipulation intent model from human video (left) and a robot controller policy from simulation (right), and pair them in real-world deployment on a dexterous hand and a parallel-jaw gripper (center).}
  \label{fig:teaser}
\end{figure}

\begin{abstract}
  The most widely-adopted robot learning pipelines today learn skills from robot demonstrations or structured human data, which are expensive to collect and tied to specific embodiments. In contrast, unstructured human videos provide a scalable alternative. They contain diverse manipulation demonstrations across objects, scenes, and strategies, but are not directly connected to robot action. We propose \pname{}, a two-stage framework that learns task intent from unstructured human videos drawn from internet-scale datasets and learns robot control in massively-parallel simulation. The intent model predicts short-horizon intent (what should happen next in the scene) from the current observation in closed loop. An embodiment-specific sensorimotor policy converts this intent into robot actions. The intent interface is shared across controllers, so the same intent model can be applied to different embodiments, from our primary dexterous hand to a parallel-jaw gripper. We evaluate \pname{} on five real-world manipulation tasks: stirring, wiping, and binning supervised by only internet video, with zero-shot transfer to novel scenes and object instances; and push-T and cable routing supervised by 1\,hr each of self-collected smartphone video. Project page: \url{https://lucid-robot.github.io/}.
\end{abstract}


\section{Introduction}

Robot manipulation policies are typically trained on action-observation data, sourced either from teleoperated robot demonstrations~\citep{zhao2023aloha,openx2024} or from human demonstrations captured with structured rigs or interfaces (motion capture, multi-view, wearables, handheld grippers)~\citep{wang2024dexcap,guzey2025aina,chi2024umi,gupta2025umionair}. These pipelines require purpose-built collection infrastructure, remain tied to a particular robot or interface, and scale only with operator hours. Two alternative sources scale beyond these limits. First, unstructured human video (internet videos) is abundant and broad in objects and strategies, but actionless. Second, physical simulation produces action-labeled data at arbitrary scale, but each task needs hand-designed rewards, especially hard to define for high-level intent. We argue these two sources are naturally complementary and should be paired through an intent-control separation: human videos provide embodiment-agnostic intent~\citep{xu2024flow,li2025novaflow}, while massive simulation provides a task-agnostic and robust sensorimotor policy~\citep{kedia2026simtoolreal,singh2024dextrahrgb}.

Prior attempts to draw on these sources each have a distinct shortcoming. Imitation policies trained on extracted human-video trajectories learn trajectory-level behavior rather than task-level intent, and don't generalize beyond the demonstrated scene~\citep{qin2022dexmv,gupta2026grasptoact}. Open-loop planners that condition a pretrained video model on the initial scene cannot recover when execution diverges~\citep{xu2024flow,kuang2026dex4d}. Pretrained video models repurposed as policy backbones still need per-task and per-embodiment robot data~\citep{gao2026dreamdojo,ye2026dreamzero}. On the simulation side, generalist sensorimotor policies generalize across objects under massive randomization~\citep{liu2025dextrack,xu2025dexplore,kedia2026simtoolreal}, but their inference-time reference still comes from outside the policy (motion capture, single-video extraction, or a video model run at task start).

We propose \pname{}, which pairs intent (what should change in the scene) and control (how the robot achieves it) through two design choices. First, intent and control are decoupled: an intent model $f_\theta$ trained on unstructured human video predicts short-horizon object flow and a palm-pose reference from the current scene, while a sensorimotor policy $\pi$ trained once in simulation realizes these references on a dexterous hand-arm system, and a parallel-jaw gripper variant demonstrates embodiment transfer. Second, intent is predicted in closed loop: at deploy, $f_\theta$ continually re-queries from the live scene rather than producing a one-shot plan, with no object mesh, motion capture, or per-embodiment adaptation required.

We evaluate \pname{} against four claims: it (1) learns intent that transfers zero-shot to novel scenes, camera viewpoints, and object instances, (2) improves robustness via closed-loop intent prediction, (3) extends across robot embodiments with the same intent model, and (4) improves predictably with the amount of training video, both in intent loss and downstream task success. We test these on five real-world manipulation tasks spanning web-scale video (stirring, wiping, binning) and self-collected smartphone video (push-T, cable routing), on a dexterous robot hand~\citep{shaw2023leaphand} and a parallel-jaw gripper. Closed-loop \pname{} achieves 73\% average success on the web-supervised tasks vs 28\% for an open-loop baseline, and the same intent model drives both embodiments with comparable success (63\%) on the smartphone-collected tasks.

\section{Related Work}

\subsection{Dexterous robot manipulation from human video}

\textbf{Trajectories from video.} A body of work trains policies directly on trajectories derived from humans performing the task, using uninstrumented internet video reconstructed or edited into 3D hand and object trajectories~\citep{qin2022dexmv,chen2025vividex,hsieh2025dexman,mu2026deximit,chen2026videomanip,gupta2026grasptoact,lum2025human2sim2robot,pan2025spider,shi2025zeromimic,lepert2025phantom} or in-scene video captured with wearables or calibrated cameras~\citep{guzey2025aina,wang2026humanego,haldar2025pointpolicy,guzey2024hudor}. These policies are scoped to specific recorded trajectories and limited by reconstruction noise and the human-to-robot embodiment gap. \pname{} uses human video only for intent supervision.

\textbf{Plans from video.} A second line uses pretrained video generators, often off-the-shelf, to produce a plan at the start of a rollout from the initial observation~\citep{xu2024flow,kuang2026dex4d,bharadhwaj2024track2act,liang2024dreamitate,li2025novaflow}. A separate executor realizes the plan, and the plan itself is generated once at $t=0$ and not updated during execution. This open-loop framing reflects the cost of video generation~\citep{liang2024dreamitate,bharadhwaj2024track2act}. \pname{} instead issues new short-horizon references continually from the current observation.

\textbf{Representations from video.} A third line treats human video as a pretraining supplement for systems that still require robot teleoperation or real-world demonstrations, whether as a pretrained video model used as a runtime component~\citep{gao2026dreamdojo,ye2026dreamzero,pai2025mimicvideo,goswami2025dexwm,routray2026vipra} or as pretraining for a VLA or transformer backbone that is then co-trained or fine-tuned with robot demonstrations~\citep{luo2025beingh0,qiu2025hat,yang2025egovla,kareer2024egomimic,li2026vitra, lepert2025masquerade, ren2025motiontracks, collins2025amplify}. Across these variants, robot action data remains the bottleneck. \pname avoids teleoperation entirely, supervising intent from human video and control from simulation.

\subsection{Sim-to-real RL for dexterous robot control}

Reinforcement learning in simulation is commonly used to train dexterous robot policies against reference hand and object trajectories from motion capture or video~\citep{liu2024quasisim,dasari2023pgdm,li2025maniptrans,mandi2025dexmachina,gupta2026grasptoact,lum2025human2sim2robot}, with each policy tied to the reference it was trained against. Recent work has scaled this lineage into generalist sensorimotor policies that span many trajectories or object instances under aggressive domain randomization, generalizing across objects, geometries, and scene conditions~\citep{liu2025dextrack,xu2025dexplore,kedia2026simtoolreal,yin2025dexgen}. References at inference still come from outside the policy (MoCap, teleop, single-video poses, or mesh+goal pose). \pname{} pairs a generalist sensorimotor policy with an intent model that produces references in closed loop.

\subsection{Intent representations for robot policies}

Human video reveals intent but not actions, so prior work uses intermediate representations to bridge the two. Object-centric flow is one such representation: point tracks or per-point trajectories extracted from human video or video generators~\citep{wen2024atm,yuan2024generalflow,ren2025motiontracks,seita2022toolflownet,xu2024flow,bharadhwaj2024track2act,li2025novaflow,kuang2026dex4d,zheng2025tracevla,huang2026pointworld}. Flow is mesh-free, covers rigid, articulated, and deformable objects, and readily extracted from internet video. But it does not specify where the hand should be at contact, an underspecification that matters for multi-fingered manipulation. Hand-centric representations (hand poses, wrist trajectories, grasp priors) carry the complementary functional grasp that flow cannot capture~\citep{mandikal2022dexvip,shi2025zeromimic,qin2022dexmv,haldar2025pointpolicy,dasari2023pgdm,li2025maniptrans,chen2025lvp}. Executors are typically lightweight (inverse kinematics, trajectory optimization, residuals on heuristic policies)~\citep{bharadhwaj2024track2act,li2025novaflow,liang2024dreamitate}, and are often the primary failure mode. \pname{} pairs flow and palm-pose intent with a generalist sensorimotor policy.

\section{Method}
\label{sec:method}

This section details \pname{}'s two learned components: the intent model $f_\theta$ (\S\ref{sec:intent}) and the sensorimotor policy $\pi$ (\S\ref{sec:policy}), which communicate through a short-horizon reference $\mathcal{R}$ comprising object flow and a palm pose. \S\ref{sec:deploy} describes how they run in closed loop at deploy.

\subsection{Intent Model}
\label{sec:intent}

\begin{wrapfigure}{R}{0.45\linewidth}
  \vspace{-1em}
  \centering
  \includegraphics[width=\linewidth]{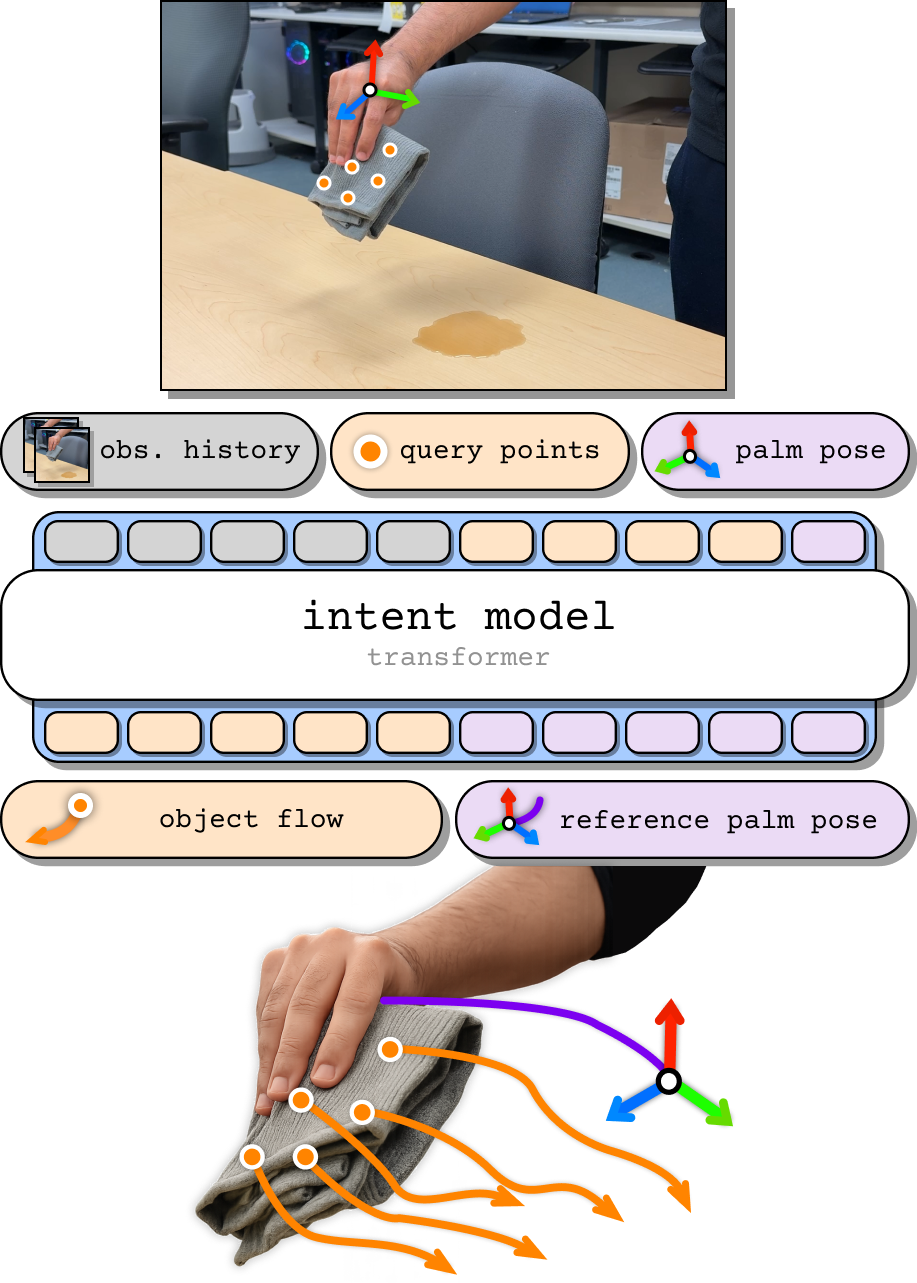}
  \caption{\textbf{Intent model.} From the recent observation history, current query points on the object, and the current palm pose, the intent model predicts short-horizon object flow and a reference palm-pose trajectory.}
  \label{fig:intent}
  \vspace{-1em}
\end{wrapfigure}

We define manipulation intent as a short-horizon prediction of object motion and rough palm pose, shared across embodiments; the joint-level commands that realize it are delegated to a separate sensorimotor policy. We predict short-horizon object and hand motion from a current RGB-D observation (Fig.~\ref{fig:intent}). Let $\mathbf{I}_t$ denote a stack of $F$ recent RGB-D frames ending at time $t$, and let $\tau=0,1,\ldots,T$ index the current step and $T$ future steps. For query point $n$ on the target object, $\mathbf{x}^{\text{trk}}_{n,\tau}\in\mathbb{R}^3$ is its 3D position at step $\tau$, and $(\mathbf{p}^{\text{palm}}_\tau, \mathbf{R}^{\text{palm}}_\tau)\in SE(3)$ is the palm pose at step $\tau$. Given $\mathbf{I}_t$ and the current-step values of these quantities, the intent model $f_\theta$ predicts the future-step values, which stack into the short-horizon reference $\mathcal{R}$ consumed by the sensorimotor policy (\S\ref{sec:policy}):
\begin{equation*}
  f_\theta\big(\mathbf{I}_t,\; \{\mathbf{x}^{\text{trk}}_{n,0}\}_{n=1}^{N},\; (\mathbf{p}^{\text{palm}}_0, \mathbf{R}^{\text{palm}}_0)\big) \;=\; \mathcal{R} \;=\; \big(\{\mathbf{x}^{\text{trk}}_{n,\tau}\}_{n=1,\tau=1}^{N,T},\; \{(\mathbf{p}^{\text{palm}}_\tau, \mathbf{R}^{\text{palm}}_\tau)\}_{\tau=1}^{T}\big).
\end{equation*}

\textbf{Architecture.} The intent model $f_\theta$ adapts CoTracker3~\citep{karaev2024cotracker3} as a point-token transformer for short-horizon prediction. We make three changes. (1) We condition the transformer on frozen DINOv3~\citep{simeoni2025dinov3} patch tokens and fuse depth through a residual adapter added to the patch-token space. (2) We predict forward in time from $\mathbf{I}_t$, whereas standard point trackers estimate tracks over already-observed frames. (3) We attach a single palm-pose token alongside the $N$ object tokens, so object flow and palm pose are produced jointly. See App.~\ref{app:intent-arch} for the full configuration.

\textbf{Training.} The intent model $f_\theta$ is trained with mean-squared-error losses on the flow ($\mathcal{L}_{\text{trk}}$), palm position ($\mathcal{L}_{\text{palm,p}}$), and palm rotation ($\mathcal{L}_{\text{palm,r}}$) streams, averaged over the $T$ future steps and combined as $\mathcal{L}(\theta) = \mathcal{L}_{\text{trk}} + \mathcal{L}_{\text{palm,p}} + \mathcal{L}_{\text{palm,r}}$. Alongside standard photometric and geometric augmentations, we heavily augment the human pixels in $\mathbf{I}_t$ (App.~\ref{app:intent-arch-aug}) so that $f_\theta$ infers object motion without leaning on demonstrator hand appearance, which matters at deploy because the visible hand is the robot, not a human.

\textbf{Supervision from unstructured video.} Per task we mine $20$\textnormal{k} clips from public video datasets~\citep{chen2024panda70m,chen2026action100m,goyal2017ssv2,damen2022epickitchens,chen2025lvp} (predominantly in-the-wild YouTube footage), or self-record $\sim$100 smartphone demonstrations for underrepresented tasks. Each clip is cut into overlapping sliding windows spanning the $F$ past frames (stacked into $\mathbf{I}_t$, ending at $\tau{=}0$) and $T$ future frames ($\tau{=}1,\ldots,T$) providing hindsight supervision for $f_\theta$. A four-stage extraction pipeline runs on each window, producing per-frame outputs indexed by $\tau$. ViPE~\citep{huang2025vipe} reconstructs per-frame camera intrinsics, extrinsics, and metric depth $D_\tau$. SAM 3.1~\citep{carion2025sam3} produces an object mask $M^{\text{obj}}_\tau$ and a human mask $M^{\text{hum}}_\tau$. DenseTrack3Dv2~\citep{ngo2025deltav2} samples $N$ query pixels inside $M^{\text{obj}}_0$ and tracks them forward; projection through $D_\tau$ yields the 3D trajectory $\{\mathbf{x}^{\text{trk}}_{n,\tau}\}$. WiLoR~\citep{potamias2024wilor} returns a per-frame MANO~\citep{romero2017mano} hand mesh, which we align to $D_\tau$ inside $M^{\text{hum}}_\tau$ via a per-frame rigid fit, yielding the palm pose $(\mathbf{p}^{\text{palm}}_\tau, \mathbf{R}^{\text{palm}}_\tau)$. The future-step slice ($\tau{=}1,\ldots,T$) of the object flow and palm pose forms the reference $\mathcal{R}$ that supervises $f_\theta$. See App.~\ref{app:supervision} for dataset details and the palm-pose lifting procedure.

\subsection{Generalist Sensorimotor Policy}
\label{sec:policy}

The sensorimotor policy $\pi$ maps onboard sensing to motor commands, and is trained via goal-conditioned RL in Isaac Lab~\citep{mittal2025isaaclab} with massively-parallel simulation to realize the intent reference $\mathcal{R}$ on a specific embodiment. We train a separate $\pi$ per embodiment with the same recipe; this section describes the dexterous-hand setup, with parallel-jaw tweaks in App.~\ref{app:baselines}.

\textbf{Setup and training data.} We train $\pi$ to follow references in $\mathcal{R}$'s format, generated procedurally rather than from $f_\theta$ or human video. Each episode loads a procedurally generated object (a union of primitives spanning blob-, tool-, and plate-like shapes) under randomized scale, mass, and friction. It then samples a reference trajectory $\mathcal{R}$ of object flow and palm pose: a chain of four segments (approach, in-hand motion, goal, disengage) with randomly sampled grasp, waypoint, and goal poses, exercising grasping, in-hand manipulation, placing, and releasing in one rollout. A per-segment hand-coupling mode varies how tightly the palm tracks the object, so the same object motion admits several valid hand strategies. Training across this broad distribution of objects and motions yields a single task-agnostic policy that tracks whatever reference $f_\theta$ produces at deploy. Geometry and sampling details are in App.~\ref{app:procedural}.

\textbf{Action space.} At each step the policy commands the arm and controls the hand through a structured grasp representation. Formally, $\pi$ outputs $\mathbf{a}_t = [\mathbf{a}_t^{\text{arm}}; \mathbf{a}_t^{\text{eig}}; \mathbf{a}_t^{\text{hnd}}]$: arm joint-position deltas $\mathbf{a}_t^{\text{arm}}$, eigen-grasp coefficients $\mathbf{a}_t^{\text{eig}}$, and per-joint hand residuals $\mathbf{a}_t^{\text{hnd}}$. The basis is fit to retargeted human grasps~\citep{ciocarlie2007eigengrasps,lum2025human2sim2robot}, so $\mathbf{a}_t^{\text{eig}}$ moves the fingers along coordinated modes of natural grasping that bias exploration toward stable grasps, while $\mathbf{a}_t^{\text{hnd}}$ adds per-joint motions the basis cannot express. The action is integrated on top of the previous joint target $\tilde{\mathbf{q}}_{t-1}$, EMA-smoothed, and clipped to joint limits to yield the new target $\tilde{\mathbf{q}}_t$ for the low-level PD controller, following SimToolReal~\citep{kedia2026simtoolreal}. See App.~\ref{app:action}.

\textbf{Observations and goal.} Both policies receive $\mathcal{R}$ as their goal alongside their state observations. Teacher $\pi^{\text{T}}$ is trained on privileged simulator state and student $\pi^{\text{S}}$ on onboard sensing; both observe the palm-pose component $\{(\mathbf{p}^{\text{palm}}_\tau, \mathbf{R}^{\text{palm}}_\tau)\}_{\tau=0}^{T}$ of $\mathcal{R}$ together with short histories of the joint configuration $\mathbf{q}_t$ and the previous joint target $\tilde{\mathbf{q}}_{t-1}$. They differ in how the object-flow component of $\mathcal{R}$ is sampled. $\pi^{\text{T}}$ sees a privileged sampling $\{\bar{\mathbf{x}}^{\text{trk}}_{n,\tau}\}_{n=1,\tau=0}^{\bar{N},T}$ drawn from the full object surface, together with joint velocities $\dot{\mathbf{q}}_t$ and the object pose. $\pi^{\text{S}}$ sees a smaller sampling $\{\mathbf{x}^{\text{trk}}_{n,\tau}\}_{n=1,\tau=0}^{N,T}$ restricted to the surface visible from an external RGB-D camera, together with a wrist-mounted depth image $\mathbf{D}^{\text{wrist}}_t$ that resolves close-range object geometry. The external camera location is randomized per episode during training to cover the viewpoints of the human videos used to train $f_\theta$. All inputs are in the robot base frame; the full table is in App.~\ref{app:observations}.

\textbf{Reward.} The sensorimotor policy $\pi$ is trained with a single task-agnostic reward, balancing three concerns. (1) Object tracking, the primary signal, rewards $\pi$ for keeping the current query-point positions $\{\bar{\mathbf{x}}^{\text{trk}}_{n,0}\}$ close to the lookahead targets $\{\bar{\mathbf{x}}^{\text{trk}}_{n,\tau}\}_{\tau=1}^{T}$, with a success bonus on reaching the goal pose. (2) Palm-pose following and finger contact enter as shaping rewards, leaving the policy room to deviate where its embodiment requires. (3) Regularization penalizes large action magnitudes $\|\mathbf{a}_t\|$ and unsafe joint configurations. The full inventory, gates, and weights are in App.~\ref{app:reward}.

\begin{figure}[t]
  \centering
  \includegraphics[width=\linewidth]{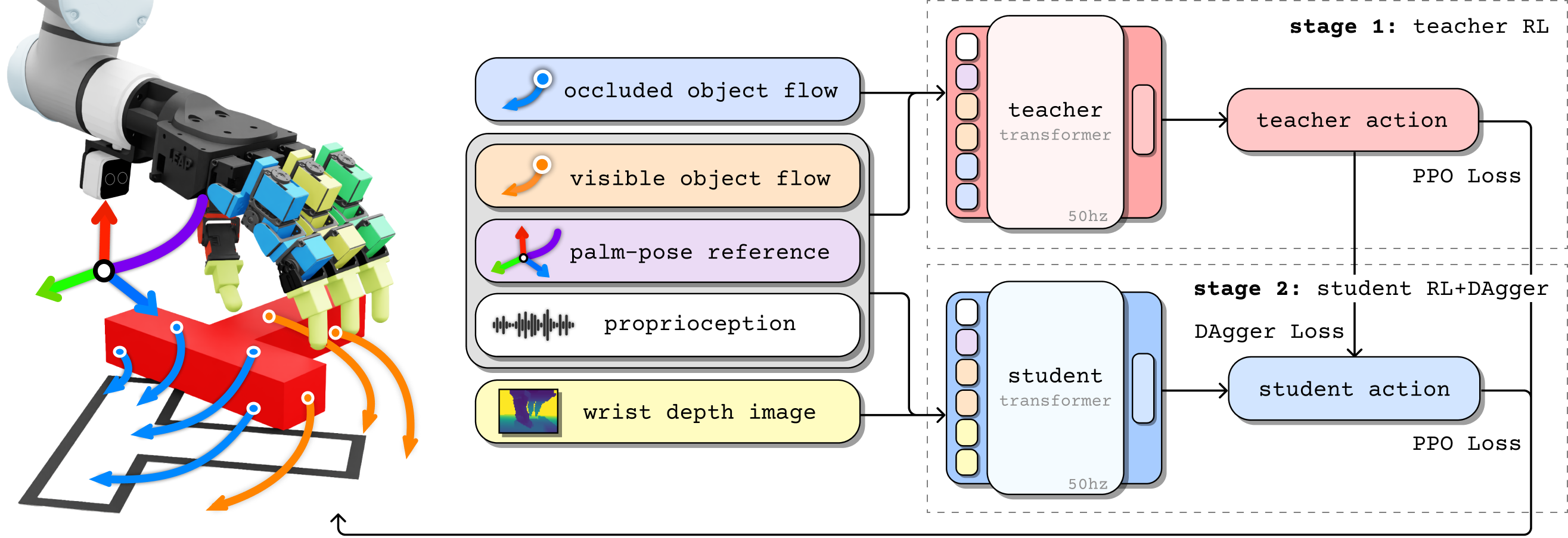}
  \caption{\textbf{Sensorimotor policy training.} The teacher $\pi^{\text{T}}$ is first trained with PPO on a privileged sampling of the object-flow component of $\mathcal{R}$ (drawn from the full object surface), the palm-pose reference, and proprioception. The student $\pi^{\text{S}}$ is then distilled from $\pi^{\text{T}}$ with a hybrid PPO + distillation objective, replacing the privileged sampling with the external camera-visible subset of the object flow plus a wrist-mounted depth image.}
  \label{fig:distillation}
  \vspace{-1em}
\end{figure}

\textbf{Policy architecture.} Both $\pi^{\text{T}}$ and $\pi^{\text{S}}$ build on AME~\citep{he2025ame}, a cross-attention architecture from legged locomotion where a proprioception-conditioned query reads from scene tokens. We adapt it in three ways. (1) The scene tokens come from query points on the object rather than an elevation grid; each query point $n$ becomes one token that encodes its full trajectory $\{\mathbf{x}^{\text{trk}}_{n,\tau}\}_{\tau=0}^{T}$ through a shared point-wise MLP, making the encoder permutation-invariant. (2) $\pi^{\text{S}}$ tokenizes the wrist-depth image $\mathbf{D}^{\text{wrist}}_t$ with a strided-conv stem and joins the resulting patch tokens to the point-trajectory tokens. (3) A self-attention block over the joint token set runs before the cross-attention, letting points and (for $\pi^{\text{S}}$) depth patches mix locally. The cross-attention output passes through an MLP trunk with proprioception to produce the action distribution and value estimate (Fig.~\ref{fig:distillation}, App.~\ref{app:teacher}).

\textbf{Training and sim-to-real.} The teacher $\pi^{\text{T}}$ is trained with PPO~\citep{schulman2017ppo} on privileged simulator state under a curriculum that tightens the environment as $\pi^{\text{T}}$ improves. As the curriculum advances, gravity rises from near-zero to full, random object wrench perturbations increase, and success tolerances on object and palm pose tighten. Each environment holds a per-episode difficulty level that rises when $\pi^{\text{T}}$ completes its trajectory within the success tolerance and falls otherwise. The population mean of these levels drives a single scalar $\rho\in[0,1]$ that interpolates these knobs, following DextrAH-RGB~\citep{akkaya2019adr,singh2024dextrahrgb} (App.~\ref{app:adr}). We also match the simulator's dynamics to the real robot via system identification on the joints (App.~\ref{app:realtosim}).

Following PHP~\citep{wu2026php}, we distill $\pi^{\text{T}}$ into the student policy $\pi^{\text{S}}$ with a hybrid objective $\mathcal{L}_{\text{student}} = \mathcal{L}_{\text{PPO}}(\pi^{\text{S}}) + \lambda_D\,\lVert \mu^{\text{S}} - \mu^{\text{T}}\rVert_2^2$, combining a PPO surrogate on on-policy rollouts of $\pi^{\text{S}}$ with an MSE regression between the teacher and student action means. The distillation weight $\lambda_D$ is annealed during training, so imitation dominates early and the on-policy PPO term takes over once $\pi^{\text{S}}$ has closed most of the gap. Throughout distillation the simulator is held at the final curriculum level $\rho=1$. The student additionally trains under realistic sensor noise on the depth cameras~\citep{handa2014benchmark} and proprioception, matching the deploy sensors. See App.~\ref{app:student} for the full loss, student network, and training schedule.

\subsection{Real-World Execution}
\label{sec:deploy}

\begin{figure}[t]
  \centering
  \includegraphics[width=\linewidth]{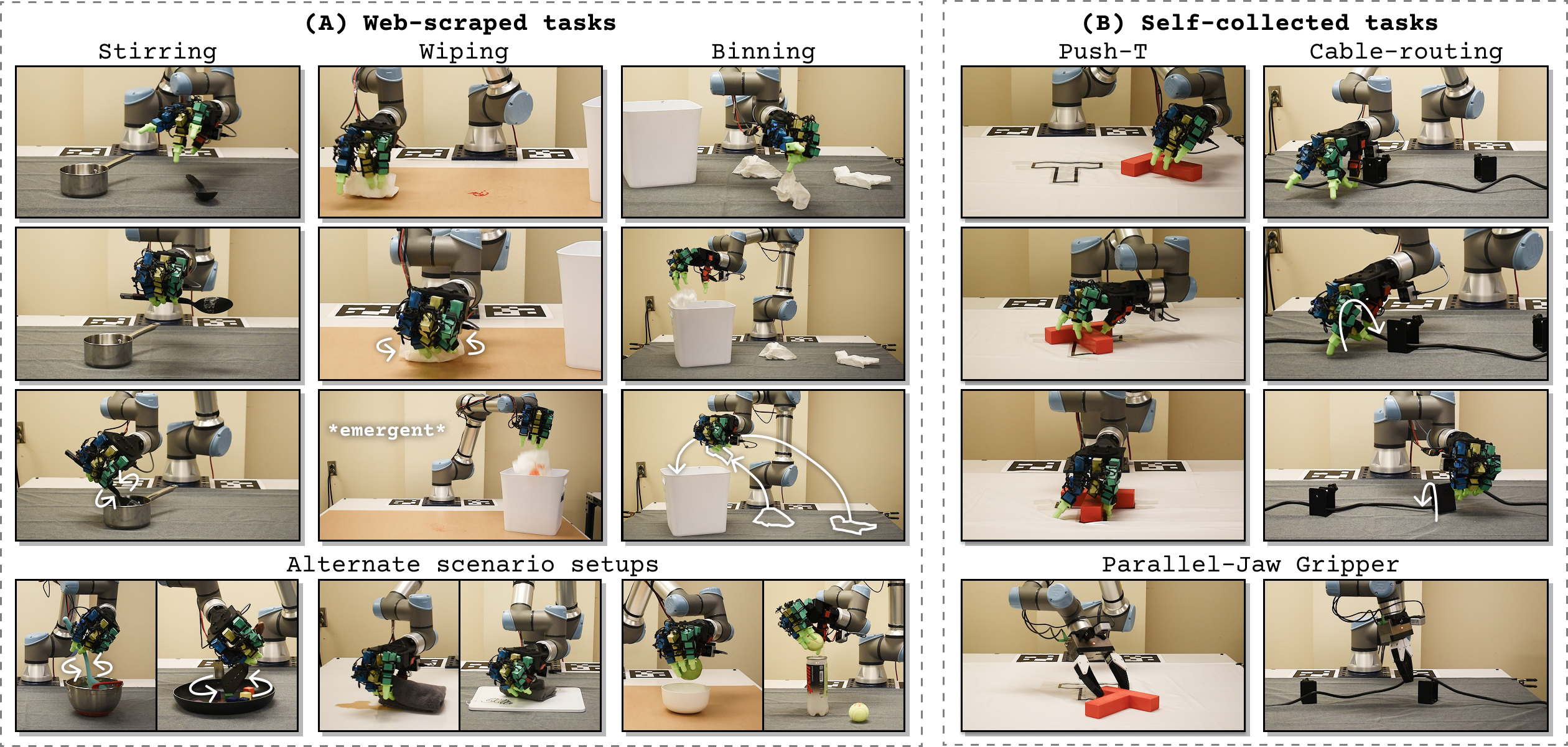}
  \caption{\textbf{Real-world tasks} we evaluated: \textbf{(A)} Three web-scraped tasks (stirring, wiping, binning), each evaluated under three scenarios. The third wiping panel shows the model depositing the used tissue in a bin without explicit binning supervision. \textbf{(B)} Two self-collected tasks (push-T, cable routing), extended to a parallel-jaw gripper setup.}
  \label{fig:real-world-tasks}
  \vspace{-1em}
\end{figure}

The robot observes through a fixed external RGB-D camera and a wrist-mounted depth camera, running a slow intent cycle around a fast control loop. On each intent cycle (mirroring the training supervision pipeline of \S\ref{sec:intent}), SAM~3.1~\citep{carion2025sam3} refreshes the object mask, query points are sampled and back-projected to current positions $\{\mathbf{x}^{\text{trk}}_{n,0}\}$, and $f_\theta$ is re-queried with these, $\mathbf{I}_t$, and the palm pose $(\mathbf{p}^{\text{palm}}_0,\mathbf{R}^{\text{palm}}_0)$ from forward kinematics to produce a fresh $\mathcal{R}$; since $f_\theta$ is trained on human video, its outputs arrive in the external-camera frame and are transformed into the robot base frame before use. At every policy step, $\pi^{\text{S}}$ consumes the current points, the flow and palm-pose reference from $\mathcal{R}$, the wrist depth $\mathbf{D}^{\text{wrist}}_t$, and proprioception, and emits $\mathbf{a}_t$. Between intent cycles a sliding-window 3D point tracker~\citep{ngo2025deltav2} keeps the current points consistent with the live scene, while a lookahead window advances over $\mathcal{R}$'s 1\,s flow horizon; once it reaches the end, we re-query $f_\theta$. Exact rates are in App.~\ref{app:deploy-pipeline}.

\section{Experimental Results}
\label{sec:exp}

We investigate four questions about \pname{}: (Q1, \S\ref{sec:exp-capability}) whether the system works end-to-end on web-scraped tasks. (Q2, \S\ref{sec:exp-scaling}) whether scaling human-video supervision improves real-world performance. (Q3, \S\ref{sec:exp-embodiment}) whether the intent model transfers across robot embodiments and supports new tasks supervised only by self-collected smartphone video. (Q4, \S\ref{sec:exp-ablations}) which tracking-policy design choices drive performance.

\subsection{Real-World Capability on Web-Scraped Tasks}
\label{sec:exp-capability}

We evaluate three web-scraped tasks, each supervised by 20k human-video clips. (1) Stirring: the robot picks up a spoon and completes three stirring circles inside the container. (2) Wiping: the robot picks up a cloth and clears random marks/substances from a surface. (3) Binning: the robot picks up each object in the workspace and deposits it in a target container. Each task is evaluated over 10 trials in each of three scenarios that jointly vary object instances, table setup, and external camera pose, all out-of-distribution for the intent model. To isolate the contribution of {\color{y2rblue}closed-loop intent}, we compare against an {\color{y2rsage}open-loop planner}: Veo~3.1~\citep{googledeepmind2026veo3modelcard} generates a single video from the initial RGB observation, whose extracted object flow and palm-pose reference drive the same sensorimotor policy (App.~\ref{app:baselines}).

\begin{figure}[t]
  \centering
  \includegraphics[width=\linewidth]{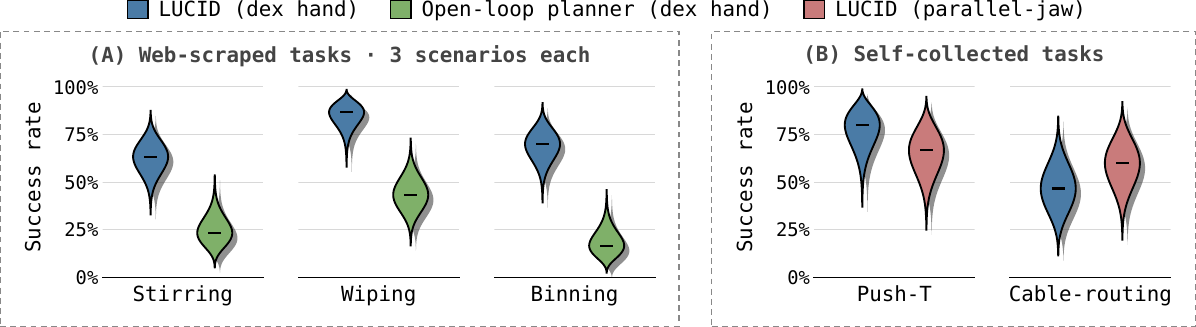}
  \caption{\textbf{Real-world success rates.} Per-task success across five real-world tasks, evaluated against task-appropriate baselines. (\textbf{A}) {\color{y2rblue}\pname{} (dex hand)} versus an {\color{y2rsage}open-loop video-generation planner (dex hand)}~\citep{googledeepmind2026veo3modelcard} on web-scale tasks. (\textbf{B}) {\color{y2rblue}\pname{} (dex hand)} versus {\color{y2rcoral}\pname{} (parallel-jaw)} on self-collected tasks. Failure-mode breakdowns appear in App.~\ref{app:per-scenario}.}
  \label{fig:success-bars}
  \vspace{-1em}
\end{figure}

\pname{}'s strength is {\color{y2rblue}closed-loop intent}: when the initial grasp misses or the object shifts mid-rollout, the intent model re-queries the scene and redirects the policy (Fig.~\ref{fig:success-bars}\textbf{A}; per-scenario in App.~\ref{app:per-scenario}). The sensorimotor policy also handles deformable objects (tissue, towel, and cloth) despite being trained only on rigid objects in simulation. We also observe emergent task composition: when we place a bin near the workspace after the robot wipes ketchup with a tissue, the intent model picks up the soiled tissue and deposits it in the bin (Fig.~\ref{fig:real-world-tasks}A, Wiping), without any task-specific binning supervision; we attribute this to incidental binning in the broader 20k-clip wiping pool. Failures cluster into perception issues from object occlusion (e.g., the spoon being covered by the hand) and unrecoverable states from unstable grasps (e.g., tennis balls slipping off the table). The {\color{y2rsage}open-loop baseline} fails differently when execution deviates from the generated plan. For example, a misgrasp during stirring sends the spoon flipping to a new location, and the stale references confuse the sensorimotor policy (App.~\ref{app:baselines}). Veo~3.1 can also hallucinate scene details, e.g., wiped regions where no wipe occurred.

\subsection{Intent Data Scaling}
\label{sec:exp-scaling}

\begin{wrapfigure}{R}{0.45\linewidth}
  \vspace{0em}
  \centering
  \includegraphics[width=\linewidth]{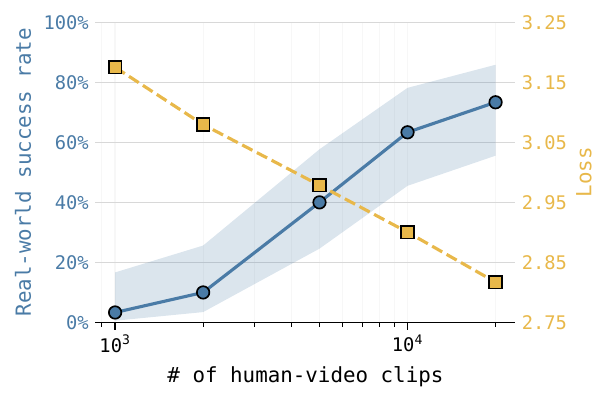}
  \caption{\textbf{Intent data scaling.} Sweeping intent-model training data from 1k to 20k human-video clips on the binning task, {\color{y2rblue}real-world success} rises and {\color{y2rmustard}held-out intent loss} falls.}
  \label{fig:scaling}
  \vspace{-1em}
\end{wrapfigure}

We sweep the binning training corpus across $\{1\textnormal{k}, 2\textnormal{k}, 5\textnormal{k}, 10\textnormal{k}, 20\textnormal{k}\}$ clips. At each scale point we run 10 real-world trials in each of the three binning scenarios from \S\ref{sec:exp-capability} (30 trials per scale point) and evaluate intent loss on a held-out set of 1k binning clips. {\color{y2rmustard}Held-out intent loss} decreases steadily with corpus size (well-fit by a power law over the observed range, see App.~\ref{app:intent-scaling-extrap}), and {\color{y2rblue}real-world success} on binning rises with it (Fig.~\ref{fig:scaling}). At 1k--2k clips the intent model produces poor references (e.g., the policy reaches the object but fails to identify the bin); in the 5k--10k range, container localization emerges but placement alignment remains weak, often missing the bin on release.

\subsection{Intent Transfer Across Robot Embodiments}
\label{sec:exp-embodiment}

To test intent transfer across embodiments, we evaluate (1) push-T~\citep{chi2023diffusionpolicy}: the robot pushes a T-shaped block to a target pose, and (2) cable routing: the robot threads a cable through two fixtures (Fig.~\ref{fig:real-world-tasks}\textbf{B}). For each task, the intent model is trained on 1\,hr of self-collected smartphone video. We deploy the same intent predictions with two sensorimotor policies trained in simulation: our primary {\color{y2rblue}LEAP hand policy}, and a {\color{y2rcoral}parallel-jaw gripper} variant with minor embodiment-specific tweaks (App.~\ref{app:baselines}). We run 15 real-world trials per embodiment per task; per-scenario breakdowns in App.~\ref{app:per-scenario}.

The {\color{y2rblue}hand} and {\color{y2rcoral}gripper} policies reach the same aggregate success on these two tasks despite very different morphologies (19/30 each; Fig.~\ref{fig:success-bars}\textbf{B}). Cable routing actually favors the gripper, since two opposing jaws are well-suited to grasping the thin cable, while the dexterous hand has a difficult time precisely grasping it. Failures on both embodiments concentrate on out-of-distribution states where the intent model's predictions fail to drive the policy forward, likely due to the small 1\,hr smartphone corpus.

\subsection{Sensorimotor Policy Ablations}
\label{sec:exp-ablations}

\begin{figure}[t]
  \centering
  \includegraphics[width=\linewidth]{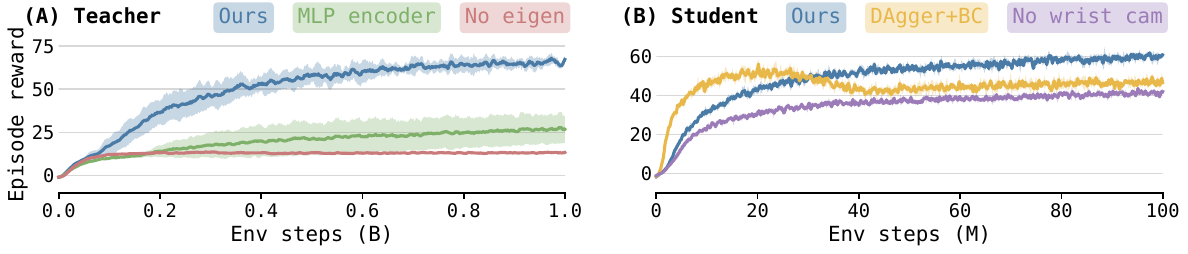}
  \caption{\textbf{Sensorimotor policy ablations.} Episode reward against environment steps for the teacher training \textbf{(A)} and the student distillation \textbf{(B)}. (\textbf{A}): {\color{y2rblue}Ours} versus an {\color{y2rsage}MLP encoder} concatenating all inputs and per-joint actions {\color{y2rcoral}without the eigen-grasp basis}. (\textbf{B}): {\color{y2rblue}Ours} versus {\color{y2rmustard}DAgger-BC distillation} and {\color{y2rlavender}no wrist camera}.}
  \label{fig:ablations}
  \vspace{-1em}
\end{figure}

Four ablations probe the choices that most shaped \pname's recipe, run in simulation with 3 seeds each. On the teacher (Fig.~\ref{fig:ablations}A) we (1) replace our policy encoder with an {\color{y2rsage}MLP encoder} concatenating all inputs without tokenization, and (2) {\color{y2rcoral}drop the eigen-grasp basis}, replacing it with per-joint actions only. On the student (Fig.~\ref{fig:ablations}B) we (3) replace our hybrid distillation objective~\cite{wu2026php} with a {\color{y2rmustard}DAgger-BC mixture}~\citep{he2025viral} over a 50/50 teacher/student rollout blend, and (4) {\color{y2rlavender}remove the wrist camera}.

The {\color{y2rsage}MLP encoder} still localizes the object but loses the per-point detail required for stable grasp contacts and in-hand manipulation. {\color{y2rcoral}Without the eigen-grasp basis} the policy faces a much larger exploration space; with the basis, training first exploits the eigen modes for natural-looking grasps and later refines individual joints for finer manipulation. {\color{y2rmustard}DAgger-BC} is a competitive baseline, but because it only mimics the teacher the student cannot fully exploit its own input modalities (e.g., the wrist camera). {\color{y2rlavender}Removing the wrist camera} leaves object geometry unresolved, similar to the MLP encoder ablation. See App.~\ref{app:query-points-ablation} for an additional ablation on query-point count $N$.

\section{Limitations}
\label{sec:limitations}
\textbf{(1) Pipeline brittleness.} The system is over-modularized; SAM~3.1, DenseTrack3Dv2, ViPE, and WiLoR each introduce a failure point during data extraction and at deployment. Perception faults like in-hand occlusion or textureless objects propagate down the chain to the sensorimotor policy. Training the intent model end-to-end from raw video would replace this chain with a single learned representation.
\textbf{(2) Task-condition gap.} Tasks without a verifiable end condition cause the intent model to loop indefinitely (stirring currently requires an externally imposed stop), and training on a new task requires manual corpus filtering, which scales poorly. Scaling to thousands of tasks through text conditioning would address both, since a high-level planner could prompt the intent model (``stir'', then ``put spoon down'') to stitch sub-tasks.
\textbf{(3) Lossy explicit interface.} The 3D-flow-plus-palm-pose interface between $f_\theta$ and $\pi^{\text{S}}$ is hand-designed and discards information such as finger configuration and fine contact. Even when intent is predicted accurately, it omits cues the policy would need to fully reproduce the human demonstration. A latent intermediate jointly optimized across both sub-models would let the interface adapt to whatever signal the policy needs~\cite{sutton2019bitter}.

\section{Conclusion}
\label{sec:conclusion}

We presented \pname{}, a framework that separates robot-agnostic intent (supervised by unstructured human video) from embodiment-specific control (learned in massively-parallel simulation). It succeeds on real-world tasks from both internet and smartphone video, scales with supervision volume, and transfers to a parallel-jaw gripper. Pairing two independently scalable supervision sources is, in our view, a promising direction for dexterous manipulation that scales without teleoperation.

\clearpage
\acknowledgments{Research supported by the NVIDIA Academic Grant Program using NVIDIA Brev. Guanya Shi holds concurrent appointments as an Assistant Professor at Carnegie Mellon University and as an Amazon Scholar. This paper describes work performed at Carnegie Mellon University and is not associated with Amazon.}

\bibliography{references}

\clearpage
\appendix

\section{Intent Model Details}
\label{app:intent}

\subsection{Supervision Pipeline}
\label{app:supervision}

This appendix details how each raw video clip is processed into the per-window supervision targets consumed by the intent-model loss (Sec.~\ref{sec:intent}).

\subsubsection{Dataset Mix and Clip Extraction}
\label{app:supervision-data}

Per task we mine $20$\textnormal{k} clips from a mix of Panda-70M~\citep{chen2024panda70m}, Action100M~\citep{chen2026action100m}, Something-Something-V2~\citep{goyal2017ssv2}, EPIC-Kitchens~\citep{damen2022epickitchens}, and the LVP~\citep{chen2025lvp} metadata release (filtered by action label and length; predominantly in-the-wild YouTube footage), or self-record on a mounted iPhone~16 Pro~Max when the task is underrepresented in those datasets. The only annotation is the object name used to prompt segmentation. Clips are resampled to $8$\,Hz and cut into sliding windows at stride $2$, following the window structure of Sec.~\ref{sec:intent}.

\subsubsection{Extraction Pipeline}
\label{app:supervision-extract}

\begin{figure}[t]
\centering
\includegraphics[width=\linewidth]{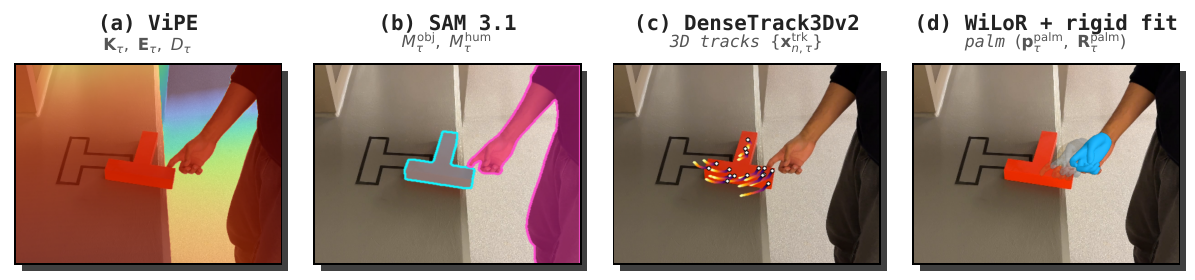}
\caption{\textbf{Supervision extraction pipeline.} Each video window is processed by four stages: \textbf{(a)}~ViPE~\citep{huang2025vipe} for camera intrinsics, extrinsics, and metric depth; \textbf{(b)}~SAM~3.1~\citep{carion2025sam3} for object and human masks; \textbf{(c)}~DenseTrack3Dv2~\citep{ngo2025deltav2} for 3D object-flow tracks; and \textbf{(d)}~WiLoR~\citep{potamias2024wilor} with a rigid fit (Eq.~\ref{eq:mano-fit}) for the palm pose. See App.~\ref{app:supervision-extract} for full details.}
\label{fig:supervision}
\end{figure}

Each window is processed by four stages, illustrated in Fig.~\ref{fig:supervision}.
\begin{enumerate}[label=\alph*),leftmargin=*]

\item \textbf{Camera and depth reconstruction.} ViPE~\citep{huang2025vipe} runs monocular SLAM and metric depth estimation, returning per-frame intrinsics $\mathbf{K}_\tau$, extrinsics $\mathbf{E}_\tau\in SE(3)$, and a dense depth map $D_\tau$.
\item \textbf{Object and hand segmentation.} SAM~3.1~\citep{carion2025sam3}, prompted with the action noun, produces a per-frame object mask $M^{\text{obj}}_\tau$ and a separate human mask $M^{\text{hum}}_\tau$. We subtract the human mask from the object mask, $M^{\text{obj}}_\tau \leftarrow M^{\text{obj}}_\tau \setminus M^{\text{hum}}_\tau$, so hand pixels never leak into the object region.
\item \textbf{Object flow tracks.} DenseTrack3Dv2~\citep{ngo2025deltav2} samples $N$ query pixels inside $M^{\text{obj}}_0$ and tracks them across the window. We back-project each tracked pixel through $D_\tau$ and $\mathbf{K}_\tau$ to obtain the 3D trajectory $\{\mathbf{x}^{\text{trk}}_{n,\tau}\}$.
\item \textbf{Hand reconstruction.} WiLoR~\citep{potamias2024wilor} returns a MANO~\citep{romero2017mano} hand mesh with wrist rotation $\mathbf{R}^{\text{W}}_\tau\in SO(3)$, in a hand-local frame at arbitrary scale. A per-frame rigid fit (Sec.~\ref{app:palm-fit}) aligns the MANO mesh to $D_\tau$ inside $M^{\text{hum}}_\tau$ and reads off the palm pose $(\mathbf{p}^{\text{palm}}_\tau, \mathbf{R}^{\text{palm}}_\tau)\in SE(3)$ at the MANO palm-center vertex.
\end{enumerate}
The intent-model supervision targets per window are the 3D query-point trajectory $\{\mathbf{x}^{\text{trk}}_{n,\tau}\}$ and the palm-pose trajectory $\{(\mathbf{p}^{\text{palm}}_\tau, \mathbf{R}^{\text{palm}}_\tau)\}$ over the $T$ future steps, which together form the reference $\mathcal{R}$ of Sec.~\ref{sec:intent}.

\subsubsection{Palm-Pose Lifting}
\label{app:palm-fit}

For each frame $\tau$ we solve a uniform scale $s\in\mathbb{R}_{>0}$ and translation $\mathbf{t}\in\mathbb{R}^3$ that align the MANO mesh with the ViPE depth at the hand pixels:
\begin{equation}
\min_{s,\mathbf{t}}\;\sum_{i\in\mathcal{V}_\tau}\bigl\lVert\,s\,\mathbf{v}^{\text{M}}_{\tau,i} + \mathbf{t} \;-\; \mathrm{unproj}\!\bigl(\mathbf{u}_i,\,\bar{d}_\tau;\,\mathbf{K}_\tau\bigr)\bigr\rVert_2^2,
\label{eq:mano-fit}
\end{equation}
where $\mathcal{V}_\tau$ is the set of back-face-culled MANO vertices filtered by $M^{\text{hum}}_\tau$, each projected to pixel $\mathbf{u}_i$, and $\bar{d}_\tau$ is the median ViPE depth over those pixels after outlier rejection. The objective is linear in $(s, t_x, t_y)$ and closed-form in $t_z$, and each frame reduces to a single \texttt{lstsq} call. We read off the palm pose as $\mathbf{p}^{\text{palm}}_\tau = s\,\mathbf{v}^{\text{palm}}_\tau + \mathbf{t}$ using the MANO palm-center vertex, and $\mathbf{R}^{\text{palm}}_\tau = \mathbf{R}^{\text{W}}_\tau\,\mathbf{R}_{\text{WM}}$ with a fixed hand-to-palm change-of-basis $\mathbf{R}_{\text{WM}}$.

\subsection{Architecture and Training Configuration}
\label{app:intent-arch}

\subsubsection{Architecture}
\label{app:intent-arch-net}

The intent model $f_\theta$ ingests a stack of $F$ past RGB-D frames and produces $T$-step predictions for $N$ object tokens and a single palm token. Core sizing is in Table~\ref{tab:intent-arch}.

\textbf{Encoder.} A frozen ViT-B/16 DINOv3~\citep{simeoni2025dinov3} encodes RGB into patch tokens. A zero-initialized residual adapter (Conv2d\,$\to$\,LayerNorm\,$\to$\,Linear) fuses depth into the same token space.

\textbf{Transformer.} The patch tokens enter an EfficientUpdateFormer from CoTracker3~\citep{karaev2024cotracker3}, running forward-in-time. Each block applies cross-attention from query tokens into the scene patch tokens, followed by spatial self-attention across the $N{+}1$ query tokens at a given future step and temporal self-attention across the $T$ future steps for each token. Each object token is initialized from its query point $\mathbf{x}^{\text{trk}}_{n,0}$ and the palm token from $(\mathbf{p}^{\text{palm}}_0, \mathbf{R}^{\text{palm}}_0)$, both encoded sinusoidally and projected to the transformer hidden size.

\textbf{Outputs.} Linear heads at each future step produce object-point displacements, palm-position displacements, and relative palm-rotation displacements. These are composed with the current query points and current palm pose to recover the future object positions and palm poses. Training minimizes mean-squared error between these outputs and the supervision trajectories of App.~\ref{app:supervision}.

\subsubsection{Training}
\label{app:intent-arch-opt}

We train a separate intent model per task. Targets are per-channel standardized using statistics from the training set, and weights are tracked by an exponential moving average used for evaluation. All training hyperparameters are in Table~\ref{tab:intent-train}. Both the supervision extraction (App.~\ref{app:supervision-extract}) and intent-model training run on a A100 GPU node.

\subsubsection{Augmentations}
\label{app:intent-arch-aug}

Standard geometric and photometric augmentations (random rotation, translation, flips, color jitter, and Gaussian noise on RGB and depth) are applied to $\mathbf{I}_t$ with probability $p_{\text{aug}}=0.8$. Human-region augmentations are applied with probability $p_{\text{human}}=0.6$ and replace the pixels inside $M^{\text{hum}}_\tau$ with one of four appearance modes drawn by a fixed mixture: heavy color jitter ($0.40$), uniform random RGB fill ($0.25$), a $2$--$6$ cell color-patch grid ($0.25$), or per-pixel random RGB ($0.10$); a brightness gradient and depth-pixel noise are applied on top.

\begin{table}[t]
\centering
\small
\caption{Intent model architecture.}
\label{tab:intent-arch}
\begin{tabular}{l c}
  \toprule
  \textbf{Hyperparameter} & \textbf{Value} \\
  \midrule
  \multicolumn{2}{l}{\emph{Inputs}} \\
  Input resolution & $256\times256$ \\
  Frame stack $F$ & $2$ \\
  Object query tokens $N$ & $16$ \\
  Future steps $T$ & $8$ \\
  Prediction horizon & $1$\,s \\
  \midrule
  \multicolumn{2}{l}{\emph{Backbone}} \\
  RGB encoder & ViT-B/16 DINOv3 (frozen) \\
  Patch size & $16\times16$ \\
  Hidden size & $768$ \\
  \midrule
  \multicolumn{2}{l}{\emph{Tracking transformer}} \\
  Backbone & EfficientUpdateFormer \\
  Depth & $12$ \\
  Hidden size & $768$ \\
  Attention heads & $12$ \\
  Factorization & spatial / temporal / scene cross-attention \\
  \midrule
  \multicolumn{2}{l}{\emph{Output heads}} \\
  Object head & Linear($768\to3$) per future step \\
  Palm head & Linear($768\to9$) per future step \\
  \bottomrule
\end{tabular}
\end{table}

\begin{table}[t]
\centering
\small
\caption{Intent model training hyperparameters.}
\label{tab:intent-train}
\begin{tabular}{l c}
  \toprule
  \textbf{Hyperparameter} & \textbf{Value} \\
  \midrule
  Optimizer & AdamW \\
  Epochs & $100$ \\
  Batch size & $16$ \\
  Learning rate & $1\times10^{-4}$ \\
  Minimum learning rate & $1\times10^{-6}$ \\
  LR schedule & cosine \\
  Warmup epochs & $5$ \\
  Weight decay & $0.01$ \\
  Gradient clip (norm) & $1.0$ \\
  EMA decay & $0.999$ \\
  \bottomrule
\end{tabular}
\end{table}

\section{Sensorimotor Policy Details}
\label{app:policy}

\subsection{Procedural objects and trajectories}
\label{app:procedural}

At environment reset we sample an object instance and a reference trajectory that specifies how the object and the palm should move through the episode.

\begin{figure}[t]
\centering
\includegraphics[width=0.9\linewidth]{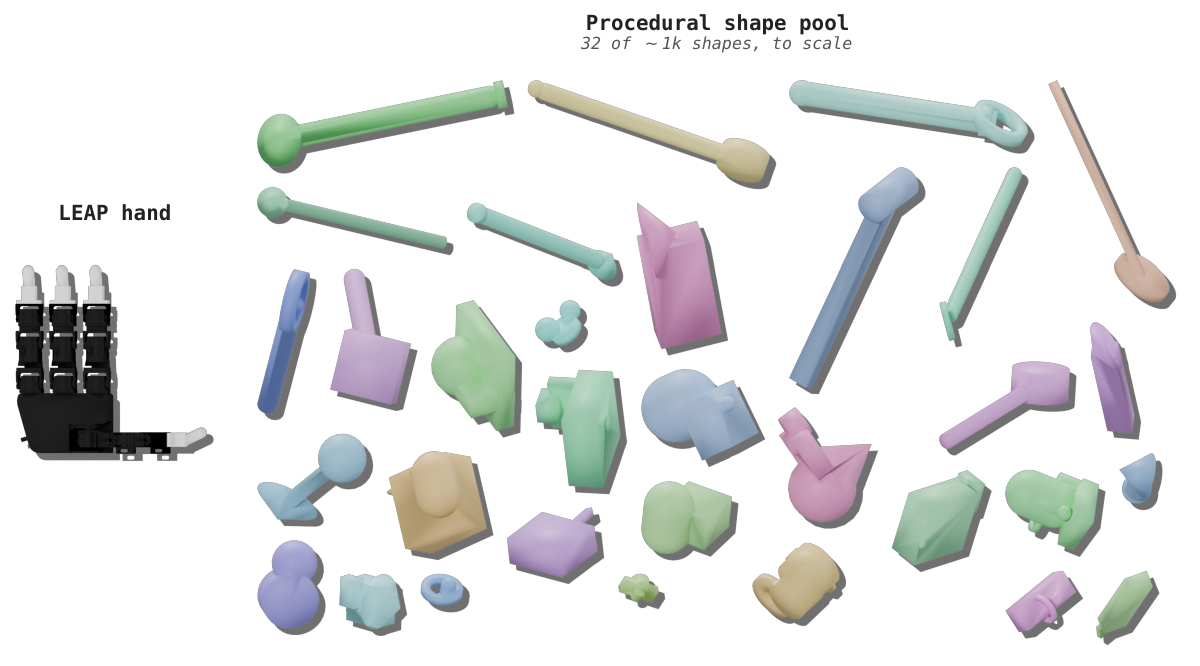}
\caption{\textbf{Procedural shape pool.} 32 random samples from the $\sim$1k-shape pool used to train $\pi$, drawn to scale beside the LEAP hand. Generation details in App.~\ref{app:procedural}.}
\label{fig:shape-pool}
\end{figure}

\textbf{Shape pool.} The object pool~(Fig.~\ref{fig:shape-pool}) is generated offline by sampling small sets of primitives (boxes, spheres, capsules, cylinders, and flat plates), attaching them at random surface points, and taking the boolean union. This covers compact blob-like, elongated tool-like, and flat book-like silhouettes. Each environment uniformly samples one asset and a scale multiplier in $[0.65, 1.1]$; mass, friction, and other properties follow App.~\ref{app:adr}.

\textbf{Reference trajectory.} A trajectory is a chain of four \emph{segments}, each defined by an object target pose, a palm target pose, a duration, and a hand-coupling mode. Within a segment, the object and palm ease smoothly from the previous segment's endpoint to the current segment's target over the assigned duration. The hand-coupling mode sets how tightly the palm tracks the object during the segment: \emph{full} rigidly locks palm position and rotation to the object, \emph{position-only} releases rotation, and \emph{none} decouples the palm entirely. The sensorimotor policy sees the next $T$ reference steps at the $8$\,Hz reference-step rate over the $1$\,s horizon emitted by $f_\theta$ at deploy. Table~\ref{tab:trajectory-segments} lists the four segments.

\begin{table}[t]
\centering
\small
\caption{Reference-trajectory segment schema. Durations in seconds.}
\label{tab:trajectory-segments}
\begin{tabular}{l l c l}
  \toprule
  \textbf{Segment} & \textbf{Object motion} & \textbf{Duration (s)} & \textbf{Hand coupling} \\
  \midrule
  Approach & static at reset pose & $2.5$ & full \\
  In-hand motion & $0$--$2$ random waypoints & $0.0$--$7.5$ & full or position-only ($0.5$/$0.5$) \\
  Goal & single waypoint & $3.0$ & full or position-only ($0.5$/$0.5$) \\
  Disengage & static at goal & $2.5$ & none \\
  \bottomrule
\end{tabular}
\end{table}

\textbf{Segment endpoint sampling.} The \emph{approach} segment ends at a grasp location sampled on the object surface, offset along the outward normal by a short standoff; contact points on the bottom half of the surface, approaches pointing toward the robot base, and upward-facing orientations are rejected. The \emph{in-hand} segment inserts $0$--$2$ random object waypoints in a bounded workspace with random per-waypoint rotations, and at each waypoint the palm-in-object frame is optionally perturbed so that the same object trajectory admits several valid hand strategies. The \emph{goal} segment ends at an object pose sampled uniformly in the workspace with random yaw, and the \emph{disengage} segment retreats the palm away from the goal.

\subsection{Action space}
\label{app:action}

The action channels $\mathbf{a}_t^{\text{arm}}\in\mathbb{R}^6$, $\mathbf{a}_t^{\text{hnd}}\in\mathbb{R}^{16}$, and $\mathbf{a}_t^{\text{eig}}\in\mathbb{R}^5$ are tanh-squashed continuous outputs.

\textbf{Eigen-grasp basis.} The basis $\mathbf{E}\in\mathbb{R}^{16\times 5}$ is the top-$5$ principal-component basis of a dataset of retargeted human grasps~\citep{lum2025human2sim2robot}, and the combined hand delta is $\Delta\mathbf{q}_t^{\text{hand}} = \mathbf{E}\mathbf{a}_t^{\text{eig}} + \mathbf{a}_t^{\text{hnd}}$.

\textbf{Integration.} Following SimToolReal~\citep{kedia2026simtoolreal}, the raw target is $\mathbf{q}_t^{\text{raw}} = \tilde{\mathbf{q}}_{t-1} + \eta\,[\mathbf{a}_t^{\text{arm}};\,\Delta\mathbf{q}_t^{\text{hand}}]$ and the PD setpoint is the EMA $\tilde{\mathbf{q}}_t = \boldsymbol{\alpha}\odot\mathbf{q}_t^{\text{raw}} + (\mathbf{1}-\boldsymbol{\alpha})\odot\tilde{\mathbf{q}}_{t-1}$ clipped to joint limits, where $\boldsymbol{\alpha}$ is a per-joint vector with $\alpha^{\text{arm}}=0.15$ on the arm block and $\alpha^{\text{hnd}}=0.75$ on the hand block, and $\eta=0.1$. At reset $\tilde{\mathbf{q}}$ is initialized to the current joint configuration so the policy begins as a residual on current state.

\subsection{Observation spaces}
\label{app:observations}

Table~\ref{tab:observations} lists the observations consumed by $\pi^{\text{T}}$ and $\pi^{\text{S}}$. $\mathbf{c}_t\in\mathbb{R}^5$ is the current hand-joint offset from the default pose projected onto $\mathbf{E}$ (App.~\ref{app:action}). The segment progress scalar is an internal clock that ramps from $0$ to $1$ between consecutive reference targets, telling the policy where it is within the current interpolation.

\begin{table}[t]
\centering
\small
\caption{Observations consumed by the teacher and student policies.}
\label{tab:observations}
\begin{tabular}{l c c c c}
  \toprule
  \textbf{Observation} & $\pi^{\text{T}}$ & $\pi^{\text{S}}$ & \textbf{Dim} & \textbf{History} \\
  \midrule
  Previous action $\mathbf{a}_{t-1}$ & \checkmark & \checkmark & $27$ & $8$ \\
  Joint positions $\mathbf{q}_t$ & \checkmark & \checkmark & $22$ & $8$ \\
  Joint velocities $\dot{\mathbf{q}}_t$ & \checkmark & & $22$ & $8$ \\
  Previous joint target $\tilde{\mathbf{q}}_{t-1}$ & \checkmark & \checkmark & $22$ & $8$ \\
  Eigen-grasp projection $\mathbf{c}_t$ & \checkmark & \checkmark & $5$ & $8$ \\
  Palm and fingertip body states & \checkmark & & $65$ & $8$ \\
  Fingertip contact forces & \checkmark & & $12$ & $8$ \\
  Mid-phalanx contact forces & \checkmark & & $9$ & $8$ \\
  Object pose (palm frame) & \checkmark & & $7$ & $8$ \\
  Object pose (world) & \checkmark & & $7$ & $8$ \\
  Object-pose lookahead & \checkmark & & $7T$ & $1$ \\
  Object-pose tracking error & \checkmark & & $7$ & $1$ \\
  Palm pose $(\mathbf{p}^{\text{palm}}_0,\mathbf{R}^{\text{palm}}_0)$ & \checkmark & \checkmark & $7$ & $8$ \\
  Palm-pose lookahead $(\mathbf{p}^{\text{palm}}_\tau,\mathbf{R}^{\text{palm}}_\tau)$ & \checkmark & \checkmark & $7T$ & $1$ \\
  Palm-pose tracking error & \checkmark & & $7$ & $1$ \\
  Full-surface query points $\bar{\mathbf{x}}^{\text{trk}}_{n,0}$ & \checkmark & & $3\bar{N}$ & $4$ \\
  Full-surface lookahead $\bar{\mathbf{x}}^{\text{trk}}_{n,\tau}$ & \checkmark & & $3\bar{N}T$ & $1$ \\
  Camera-visible query points $\mathbf{x}^{\text{trk}}_{n,0}$ & & \checkmark & $3N$ & $1$ \\
  Camera-visible lookahead $\mathbf{x}^{\text{trk}}_{n,\tau}$ & & \checkmark & $3NT$ & $1$ \\
  Segment progress scalar & \checkmark & \checkmark & $1$ & $1$ \\
  Wrist depth image $\mathbf{D}^{\text{wrist}}_t$ & & \checkmark & $80{\times}60$ & $1$ \\
  \bottomrule
\end{tabular}
\end{table}

\subsection{Reward composition}
\label{app:reward}

The sensorimotor policy is trained with a single task-agnostic reward (Table~\ref{tab:reward-terms}). Each episode is split into three phases aligned with the trajectory segments of App.~\ref{app:procedural}: $G\,{=}\,\text{approach}$, $M\,{=}\,\text{in-hand motion}+\text{goal}$, $R\,{=}\,\text{disengage}$. Throughout, $(e_p, e_\theta)$ denote object position/rotation tracking errors and $(e^h_p, e^h_\theta)$ the palm counterparts; position values are in meters and rotation values in radians, including inside the exp and tanh kernels.

Two shared gates appear in many rows. The \emph{contact factor} $c(\mathbf{f})$ gates rewards that should only pay out during a thumb-opposed grasp, and adds a small capped bonus when the palm and proximal finger links also contact the object. Using the saturation kernel $s(\cdot)$ of Table~\ref{tab:reward-symbols}, with $s_i=s(\|\mathbf{f}_i\|)$ per fingertip and $s_{\text{palm}}=s(F_{\text{palm}})$,
\begin{equation}
\begin{aligned}
  c(\mathbf{f}) &= s_{\text{thumb}}\cdot\big(2.0\,s_{(0)} + 1.8\,s_{(1)} + 1.4\,s_{(2)}\big)\;+\;0.4\,s_{\text{palm}}, \\
  s_{(k)} &\text{ is the $k$-th largest of $\{s_{\text{index}},s_{\text{middle}},s_{\text{ring}}\}$.}
\end{aligned}
\label{eq:contact-factor}
\end{equation}
The \emph{palm-pose gate} $g_{\text{hp}}$ gates tracking rewards on palm alignment,
\begin{equation}
g_{\text{hp}} = \min\!\big(r(e^h_p;\,\delta^{\text{hp}}_p),\ r(e^h_\theta;\,\delta^{\text{hp}}_\theta)\big),\qquad r(x;[a,b]){=}\operatorname{clip}((b-x)/(b-a),0,1),
\label{eq:palm-pose-gate}
\end{equation}
with ramps $\delta^{\text{hp}}_p{=}(0.04, 0.06)$ and $\delta^{\text{hp}}_\theta{=}(0.4, 0.6)$.

The remaining symbols used in Table~\ref{tab:reward-terms} are collected in Table~\ref{tab:reward-symbols}.

\begin{table}[t]
\centering
\footnotesize
\caption{Symbols used in Table~\ref{tab:reward-terms}.}
\label{tab:reward-symbols}
\begin{tabularx}{\linewidth}{l X}
  \toprule
  \textbf{Symbol} & \textbf{Meaning} \\
  \midrule
  \multicolumn{2}{l}{\emph{Phase}} \\
  $\phi\in\{G,M,R\}$                      & current phase. \\
  \midrule
  \multicolumn{2}{l}{\emph{Fingertip contact}} \\
  $s(x)$ & $\operatorname{clip}((x-0.5)/1.5,0,1)$ kernel applied to per-body force magnitudes. \\
  $\mathbf{f}_i$                           & force vector at fingertip $i\in\{\text{thumb, index, middle, ring}\}$. \\
  $F_{\text{palm}}$                   & summed force over palm and proximal-finger links. \\
  $F_{\text{tot}}=\sum_i\|\mathbf{f}_i\|$  & total fingertip force. \\
  $g_{\text{cont}}(F_{\text{tot}})$        & contact-required gate, $0.05\to 1$ as $F_{\text{tot}}$ ramps $1\to 2$\,N. \\
  $\mathbf{f}^{\text{net}}_b,\mathbf{f}^{\text{obj}}_b$ & net and object-filter contact forces on self-contact link $b\in\mathcal{B}_{\text{sc}}$. \\
  \midrule
  \multicolumn{2}{l}{\emph{Object tracking}} \\
  $e_t$                                     & scalar aggregate of object tracking error at time $t$. \\
  $e_{t-\Delta}$                            & same error one reference step earlier. \\
  $\Delta$                                  & one reference step ($8$\,Hz over the $1$\,s horizon emitted by $f_\theta$). \\
  $\Delta\mathbf{p}$                        & object-in-palm position change over $\Delta$. \\
  $\Delta\theta$                            & object-in-palm rotation change over $\Delta$. \\
  $(\sigma_p,\sigma_\theta)=(0.03,0.4)$    & palm-pose-following kernel widths (m, rad). \\
  $(\tau_p,\tau_\theta)$                    & trajectory-success thresholds, ADR $(0.07,0.7)\to(0.04,0.4)$. \\
  $\mathbf{v}_{\text{obj}}$                 & object linear velocity. \\
  $\boldsymbol{\omega}_{\text{obj}}$        & object angular velocity. \\
  \midrule
  \multicolumn{2}{l}{\emph{Robot state}} \\
  $\mathbf{q}^{\text{hand}}$                & hand sub-vector of $\mathbf{q}$. \\
  $\mathbf{q}^{\text{hand}}_{\text{def}}$   & default hand pose. \\
  $N_{\text{arm}}, N_{\text{hand}}$         & arm (6) and hand (16) DOF counts. \\
  $l$                                       & arm-link (incl. palm) index. \\
  $z_l$                                     & base-frame height of link $l$. \\
  $h_l$                                     & clearance threshold, $0.1$ mid-arm and $0.05$ distal/palm. \\
  \bottomrule
\end{tabularx}
\end{table}

\begin{table}[t]
\centering
\footnotesize
\caption{Reward terms. Phase letters $G,M,R$ map to approach, in-hand+goal, and disengage.}
\label{tab:reward-terms}
\begin{tabularx}{\linewidth}{l r l X}
  \toprule
  \textbf{Term} & \textbf{Weight} & \textbf{Phase} & \textbf{Expression} \\
  \midrule
  \multicolumn{4}{l}{\emph{Tracking}} \\
  Lookahead tracking & $+8.0$ & $G, M$ & $\tfrac12[e^{-e_p/0.03}+e^{-e_\theta/0.4}]\cdot c(\mathbf{f})\cdot g_{\text{hp}}$ \\
  Trajectory success & $+25.0$ & $R$ & $0.1\cdot\tfrac12[e^{-e_p/0.03}+e^{-e_\theta/0.4}]$ \newline \hspace*{1em}$+\;0.9\cdot\mathbb{1}[e_p{<}\tau_p \wedge e_\theta{<}\tau_\theta]\cdot g_{\text{cont}}(F_{\text{tot}})\cdot g_{\text{hp}}$ \\
  Tracking progress & $+4.0$ & all & $\operatorname{clip}\big((e_{t-\Delta}-e_t)/e_{t-\Delta},\ 0,1\big)$ \\
  Timeout bonus & $+200.0$ & all & $\mathbb{1}[\text{natural episode end}]$ \\
  \midrule
  \multicolumn{4}{l}{\emph{Palm-pose shaping}} \\
  Palm-pose following & $+3.5$ & all & $\tfrac12[e^{-e^h_p/\sigma_p}+e^{-e^h_\theta/\sigma_\theta}]$ \\
  Good finger contact & $+1.5$ & $M, R$ & $[\mathbb{1}_{\phi=R}(1-c(\mathbf{f})) + \mathbb{1}_{\phi\neq R}\,c(\mathbf{f})]\cdot g_{\text{hp}}$ \\
  Finger manipulation & $+6.0$ & $M, R$ & $\tfrac12[(1-e^{-\|\Delta\mathbf{p}\|/0.015})+(1-e^{-|\Delta\theta|/0.15})]\cdot \mathbb{1}[e_t{<}e_{t-\Delta}]\cdot c(\mathbf{f})\cdot g_{\text{hp}}$ \\
  \midrule
  \multicolumn{4}{l}{\emph{Regularizers and safety}} \\
  Action $L_2$ & $-0.05$ & all & $\tfrac{1}{N_{\text{arm}}}\|\mathbf{a}_t^{\text{arm}}\|^2+\tfrac{0.175}{N_{\text{hand}}}\,(\|\mathbf{a}_t^{\text{eig}}\|^2+\|\mathbf{a}_t^{\text{hnd}}\|^2)$ \\
  Finger regularizer & $-1.5$ & $G, R$ & $1-\exp\!\big(-\sqrt{\tfrac{1}{N_{\text{hand}}}\|\mathbf{q}^{\text{hand}}-\mathbf{q}^{\text{hand}}_{\text{def}}\|^2}\,/\,0.4\big)$ \\
  Object stillness & $-0.3$ & $R$ & $\tfrac12[\tanh(\|\mathbf{v}_{\text{obj}}\|/0.02)+\tanh(\|\boldsymbol{\omega}_{\text{obj}}\|/0.2)]$ \\
  Arm-table penalty & $-1.0$ & all & $\operatorname{mean}_{l}\mathbb{1}[z_l<0.255+h_l]$ \\
  Finger self-contact & $-0.1$ & all & $\operatorname{mean}_{b\in\mathcal{B}_{\text{sc}}}\|\mathbf{f}^{\text{net}}_b-\mathbf{f}^{\text{obj}}_b\|$ \\
  Early termination & $-100.0$ & all & $\mathbb{1}[\text{safety termination}]$ \\
  \bottomrule
\end{tabularx}
\end{table}

\textbf{Termination.} An episode terminates early and pays the termination penalty if the object or palm pose deviates from its reference beyond an ADR-interpolated threshold (App.~\ref{app:adr}) or if any joint enters a non-physical configuration. A natural termination occurs at the end of the disengage segment and pays the timeout bonus.

\subsection{Teacher and student training}
\label{app:teacher}
\label{app:student}

Widths for both networks and all PPO and distillation hyperparameters are in Tables~\ref{tab:networks} and~\ref{tab:rl-hparams}. Training uses rl\_games~\citep{makoviichuk2021rlgames}. All teacher PPO and student distillation runs use a single RTX~5090 GPU.
\begin{table}[t]
\centering
\small
\caption{Teacher and student network architectures. Widths are per-block.}
\label{tab:networks}
\begin{tabularx}{\linewidth}{@{}l >{\raggedright\arraybackslash}X@{}}
  \toprule
  \textbf{Component} & \textbf{Configuration} \\
  \midrule
  \multicolumn{2}{@{}l}{\emph{Teacher $\pi^{\text{T}}$}} \\
  Token hidden dim & $64$ \\
  Attention heads & $4$ (head dim $16$) \\
  Point encoder & per-point MLP $[64, 64]$, ELU \\
  Proprio encoder (query) & MLP $[64, 64]$, ELU \\
  Cross-attention (query $\to$ points) & $1$ block \\
  Shared trunk & MLP $[512, 256, 128]$, ELU \\
  Heads & linear $\mu\in\mathbb{R}^{27}$, $\log\sigma\in\mathbb{R}^{27}$, $V\in\mathbb{R}$ \\
  \midrule
  \multicolumn{2}{@{}l}{\emph{Student $\pi^{\text{S}}$}} \\
  Token hidden dim & $128$ \\
  Attention heads & $8$ (head dim $16$) \\
  Wrist-depth tokenizer & $4$ stride-$2$ conv stages, channels $[32, 64, 128, 128]$, BN+ReLU; flattened to patch tokens of dim $128$ \\
  Point encoder & per-point MLP $[128, 128]$, ELU \\
  Proprio encoder (query) & MLP $[128, 128]$, ELU (proprioception only) \\
  Cross-attention (query $\to$ point $\cup$ depth tokens) & $1$ block \\
  Shared trunk & MLP $[1024, 512, 256, 128]$, ELU \\
  Heads & linear $\mu\in\mathbb{R}^{27}$, $\log\sigma\in\mathbb{R}^{27}$, $V\in\mathbb{R}$ \\
  \bottomrule
\end{tabularx}
\end{table}

\textbf{Distillation loss.} The main-text objective in full, with the fixed BC weight $\kappa$ written explicitly (it is absorbed into $\lambda_D$ in the main text), is
\begin{equation}
\mathcal{L}_{\text{student}} \;=\; \mathcal{L}_{\text{PPO}}(\pi^{\text{S}}) \;+\; \lambda_D\cdot \kappa \cdot \mathbb{E}\big[\lVert \mu^{\text{S}}(o^{\text{S}}) - \mu^{\text{T}}(o^{\text{T}})\rVert_2^2\big],
\label{eq:distill-full}
\end{equation}
where $o^{\text{S}}$ and $o^{\text{T}}$ are the paired onboard and privileged observations on the same transition, $\mathcal{L}_{\text{PPO}}$ aggregates the standard clipped surrogate, critic regression, entropy bonus, and action-bounds penalty, and $\kappa$ is a fixed BC weight. $\lambda_D$ is annealed linearly from $1.0$ to $0.1$ over the first $1{,}000$ epochs. The wrist-depth image is additionally perturbed during distillation with the correlated depth-noise model~\citep{handa2014benchmark}.

\begin{table}[t]
\centering
\small
\caption{Teacher PPO and student distillation hyperparameters.}
\label{tab:rl-hparams}
\begin{tabular}{l c}
  \toprule
  \textbf{Hyperparameter} & \textbf{Value} \\
  \midrule
  \multicolumn{2}{l}{\emph{Teacher PPO}} \\
  Parallel environments & $20{,}480$ \\
  Horizon (steps / env / rollout) & $32$ \\
  Minibatch size & $32{,}768$ \\
  Mini-epochs per update & $4$ \\
  Optimizer & Adam \\
  Learning rate & $5\times10^{-4}$ (adaptive KL) \\
  KL threshold & $0.016$ \\
  Discount $\gamma$ & $0.998$ \\
  GAE $\lambda$ & $0.95$ \\
  Entropy coefficient & $0.0025$ \\
  Clip ratio & $0.2$ \\
  Critic coefficient & $4.0$ \\
  Gradient-norm clip & $1.0$ \\
  \midrule
  \multicolumn{2}{l}{\emph{Student distillation}} \\
  Parallel environments & $2{,}048$ \\
  Horizon (steps / env / rollout) & $16$ \\
  Minibatch size & $8{,}192$ \\
  Mini-epochs per update & $4$ \\
  Learning rate & $3\times10^{-4}$ (adaptive KL) \\
  Entropy coefficient & $0.002$ \\
  BC coefficient $\kappa$ & $10.0$ \\
  $\lambda_D$ initial $\to$ floor & $1.0\to 0.1$ (linear, $1{,}000$ epochs) \\
  Other settings & as teacher \\
  \bottomrule
\end{tabular}
\end{table}

\subsection{Adaptive domain randomization and curriculum}
\label{app:adr}

Domain randomization combines \emph{static DR}, sampled at reset from the fixed ranges in Table~\ref{tab:static-dr}, and \emph{adaptive DR}, which interpolates the parameters in Table~\ref{tab:adr-tandem} under a curriculum scalar $\rho\in[0,1]$. Each environment holds a per-episode difficulty $d_i\in\{0,\ldots,10\}$ that is promoted on trajectory success and demoted otherwise, and $\rho = \tfrac{1}{10 N_{\text{env}}}\sum_i d_i$ is the normalized population mean over the $N_{\text{env}}$ parallel environments. Following DextrAH-RGB~\citep{singh2024dextrahrgb,akkaya2019adr}, we bind the whole curriculum to this scalar but compute it from the per-environment population rather than a global success average.

\begin{table}[t]
\centering
\small
\caption{Static domain-randomization ranges.}
\label{tab:static-dr}
\begin{tabular}{l l l}
  \toprule
  \textbf{Group} & \textbf{Parameter} & \textbf{Range} \\
  \midrule
  Object & per-instance scale & $[0.65, 1.1]$ \\
  & mass (scale of nominal) & $[0.1, 1.6]$ \\
  & friction (static / dynamic) & $[0.5, 1.0]$ \\
  Robot body & friction (static / dynamic) & $[0.5, 1.0]$ \\
  Robot actuators & stiffness (scale of nominal) & $[0.8, 1.2]$ \\
  & damping (scale of nominal) & $[0.8, 1.2]$ \\
  & joint-friction (scale of nominal) & $[0.8, 1.2]$ \\
  \midrule
  Reset-time jitter & robot base $(x, y, z)$ & $[-0.01, 0.01]$\,m \\
  & wrist-camera forward offset & $[0.0, 0.002]$\,m \\
  & wrist-camera lateral/vertical offset & $[-0.002, 0.002]$\,m \\
  \midrule
  Student obs noise ($\sigma$) & joint positions (arm) & $0.005$\,rad \\
  & joint positions (hand) & $0.05$\,rad \\
  & eigen-grasp projection $\mathbf{c}_t$ & $0.01$ \\
  & palm pose, palm-pose lookahead & $0.005$ each \\
  & camera-visible query points, lookahead & $0.005$ each \\
  \bottomrule
\end{tabular}
\end{table}

\begin{table}[t]
\centering
\small
\caption{ADR-interpolated parameters. All sweep linearly from initial to final as $\rho: 0\to 1$.}
\label{tab:adr-tandem}
\begin{tabular}{l l l}
  \toprule
  \textbf{Parameter} & \textbf{Initial ($\rho=0$)} & \textbf{Final ($\rho=1$)} \\
  \midrule
  Gravity $\|\mathbf{g}\|$ & $0.0981$ m/s$^2$ & $9.81$ m/s$^2$ \\
  Trajectory-deviation termination $(e_p,e_\theta)$ & $(0.20$\,m, $2.0$\,rad$)$ & $(0.10$\,m, $1.0$\,rad$)$ \\
  Palm-pose termination $(e^h_p,e^h_\theta)$ & $(0.15$\,m, $2.0$\,rad$)$ & $(0.075$\,m, $0.9$\,rad$)$ \\
  Trajectory-success tolerance $(\tau_p,\tau_\theta)$ & $(0.07$\,m, $0.7$\,rad$)$ & $(0.04$\,m, $0.4$\,rad$)$ \\
  Object wrench (force, torque) per axis & $0$, $0$ & $\pm 0.15$\,N, $\pm 0.01$\,N\,m \\
  \bottomrule
\end{tabular}
\end{table}

\subsection{Real-to-sim calibration}
\label{app:realtosim}

Several measures complement the parametric randomization of Table~\ref{tab:static-dr} and the ADR schedule of Table~\ref{tab:adr-tandem}.
\begin{itemize}[leftmargin=1.2em,itemsep=2pt,topsep=2pt]
\item \textbf{Actuator system identification.} Per-joint stiffness, damping, and joint friction on the UR5e arm and LEAP hand are identified with CMA-ES~\citep{hansen2001cmaes} against recorded step responses; the fit sets the nominal actuator values that Table~\ref{tab:static-dr} perturbs.
\item \textbf{Depth rendering and noise.} Wrist and external depth streams are rendered in Isaac Lab at the native resolutions of the deployment sensors and perturbed at observation time with the correlated depth-noise model of~\citet{handa2014benchmark}.
\item \textbf{Reset-time jitter.} At each episode reset we apply static jitter to the robot's joint positions, base pose, and wrist-camera mounting offset.
\item \textbf{Student observation noise.} During distillation every student observation channel carries per-channel Gaussian noise at the fixed amplitudes listed at the bottom of Table~\ref{tab:static-dr}.
\end{itemize}

\subsection{Deployment pipeline}
\label{app:deploy-pipeline}

\begin{figure}[t]
\centering
\includegraphics[width=\linewidth]{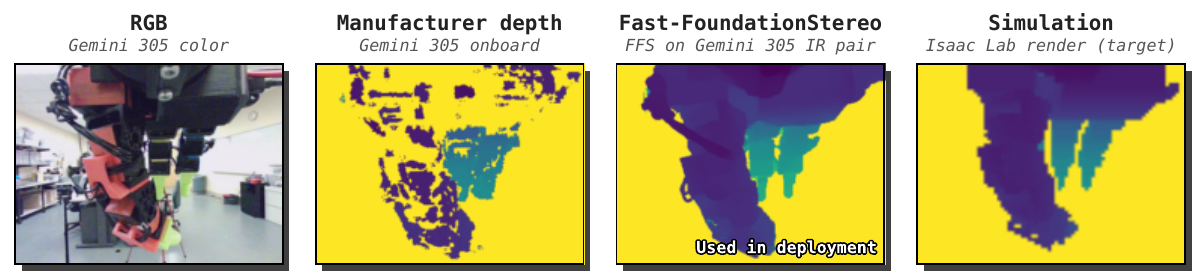}
\caption{\textbf{Wrist-camera depth.} RGB and three depth streams from the wrist-mounted Gemini 305: the manufacturer's onboard depth, Fast-FoundationStereo~\citep{wen2026fastfoundationstereo} on the IR pair (5\,ms inference), and the Isaac Lab simulation depth used at training. \pname{} deploys with the FFS stream.}
\label{fig:wrist-depth}
\end{figure}

The sensorimotor policy $\pi^{\text{S}}$ is ticked at the same $50$\,Hz as in simulation. Wrist-camera depth is produced by Fast-FoundationStereo~\citep{wen2026fastfoundationstereo} on the IR pair rather than the sensor's onboard depth stream, which we found too noisy at the close range that $\pi^{\text{S}}$ conditions on (Fig.~\ref{fig:wrist-depth}). Rates of every pipeline stage are listed in Table~\ref{tab:deploy-rates}. The entire deployment stack (SAM~3.1, intent model, point tracker, sensorimotor policy, Fast-FoundationStereo) runs on a single RTX~5090 GPU.

\begin{table}[t]
\centering
\small
\caption{Deployment tick rates.}
\label{tab:deploy-rates}
\begin{tabular}{l c}
  \toprule
  \textbf{Stage} & \textbf{Rate (Hz)} \\
  \midrule
  SAM~3.1 mask refresh                                           & 1 \\
  Intent model $f_\theta$                                        & 1 \\
  DenseTrack3Dv2 sliding-window tracking                         & 30 \\
  Sensorimotor policy $\pi^{\text{S}}$                           & 50 \\
  \bottomrule
\end{tabular}
\end{table}

\section{Experimental Details}
\label{app:experiments}

\begin{figure}[t]
\centering
\includegraphics[width=\linewidth]{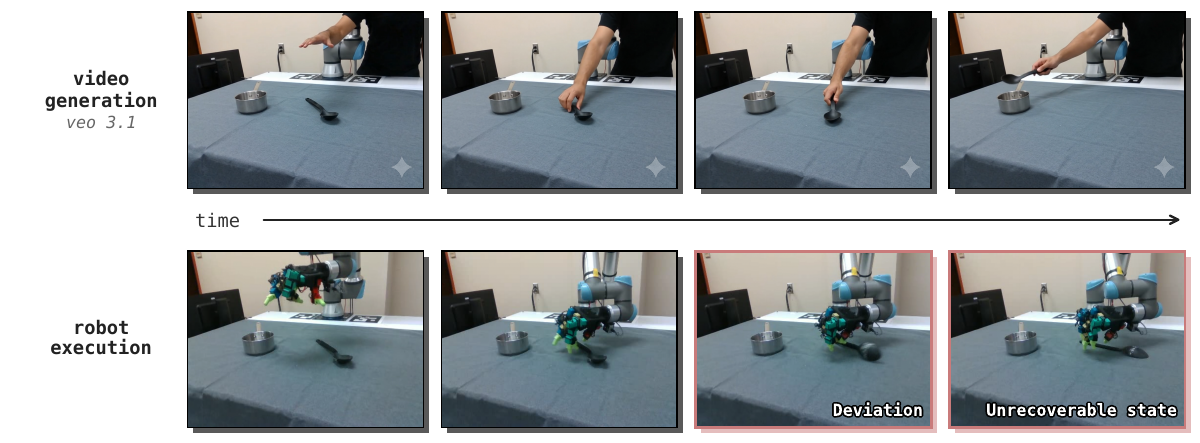}
\caption{\textbf{Open-loop video-generation planner.} Veo~3.1 generates a human video plan from the initial scene; object flow and palm pose are extracted and executed by the sensorimotor policy. The plan is fixed, so execution can diverge.}
\label{fig:open-loop-planner}
\end{figure}

\subsection{Baselines}
\label{app:baselines}

\textbf{Open-loop video-generation planner (\S\ref{sec:exp-capability}).} For each rollout (each generation takes 2--3 minutes), we generate a fresh video plan from the initial RGB observation, filtering out generations that depict random or clearly incorrect behavior and resampling until a plausible plan is obtained. Object flow and the palm-pose reference are extracted with the same pipeline used for human-video supervision (App.~\ref{app:supervision}), except that the entire generated video is processed as one continuous window rather than overlapping windows, so a single non-windowed flow and palm-pose trajectory is produced. See Fig.~\ref{fig:open-loop-planner} for an example rollout.

\textbf{Parallel-jaw gripper sensorimotor policy (\S\ref{sec:exp-embodiment}).} Trained in simulation with the same procedural trajectories, objects, and reward as the LEAP hand policy, with the \emph{Finger manipulation} reward removed. The two gripper jaws stand in for the fingers in the remaining contact-gated terms: the contact factor $c(\mathbf{f})$ collapses to $c(\mathbf{f}) = 2\,s_{\text{jaw1}} + 2\,s_{\text{jaw2}}$, and \emph{Good finger contact} uses this collapsed form. The intent model's palm-pose reference, supervised against a human palm, is mapped to the gripper by aligning the palm frame so that the gripper's closing axis matches the four-fingers-to-thumb opposition of the human hand.

\subsection{Per-scenario failure-mode breakdown}
\label{app:per-scenario}

\begin{figure}[t]
\centering
\includegraphics[width=0.99\linewidth]{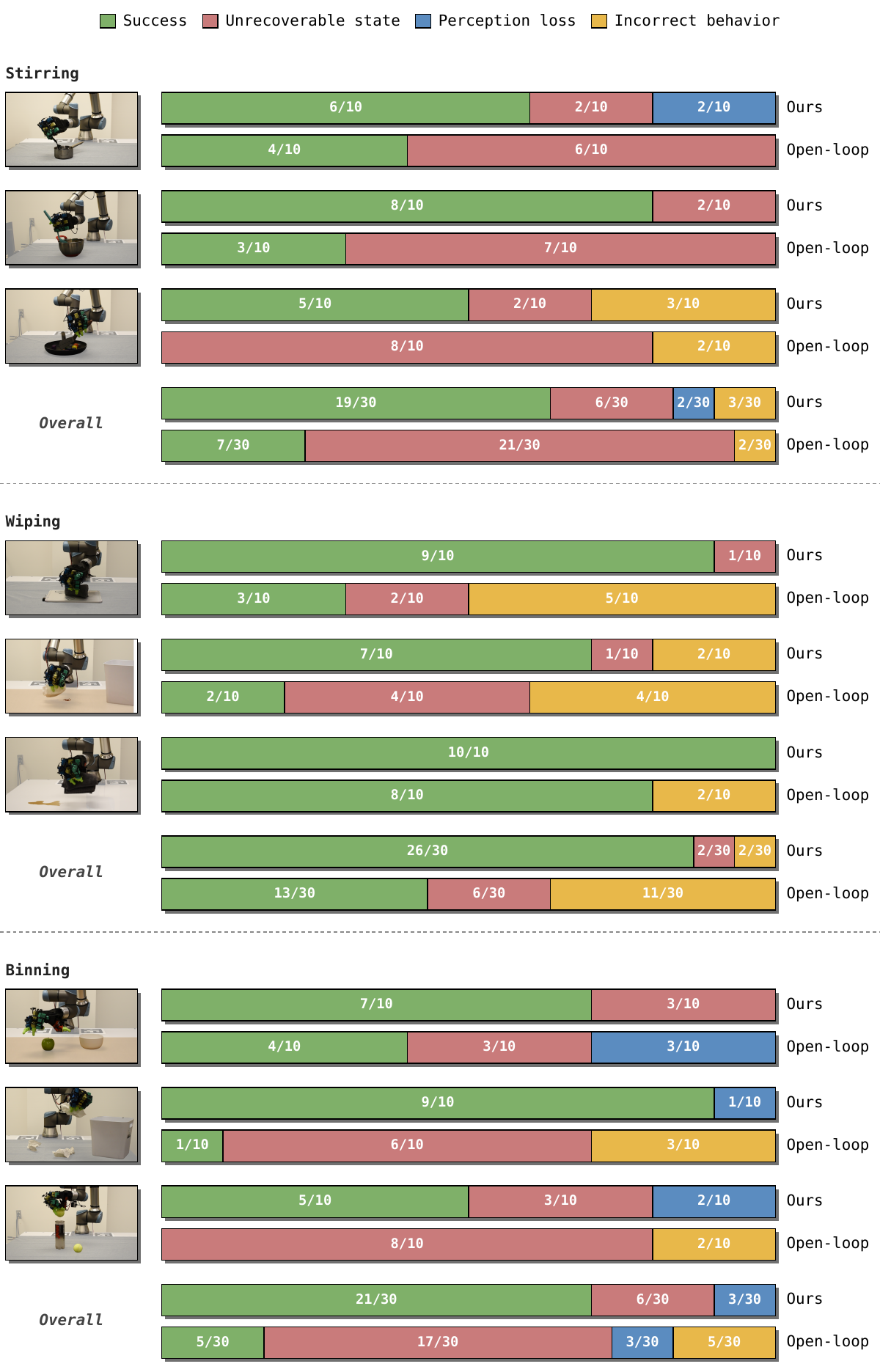}
\caption{\textbf{Failure-mode breakdown: web-scale tasks.} Per-trial outcomes for Stirring, Wiping, and Binning across the three evaluation scenarios from \S\ref{sec:exp-capability}.}
\label{fig:failures-web}
\end{figure}

\begin{figure}[t]
\centering
\includegraphics[width=0.99\linewidth]{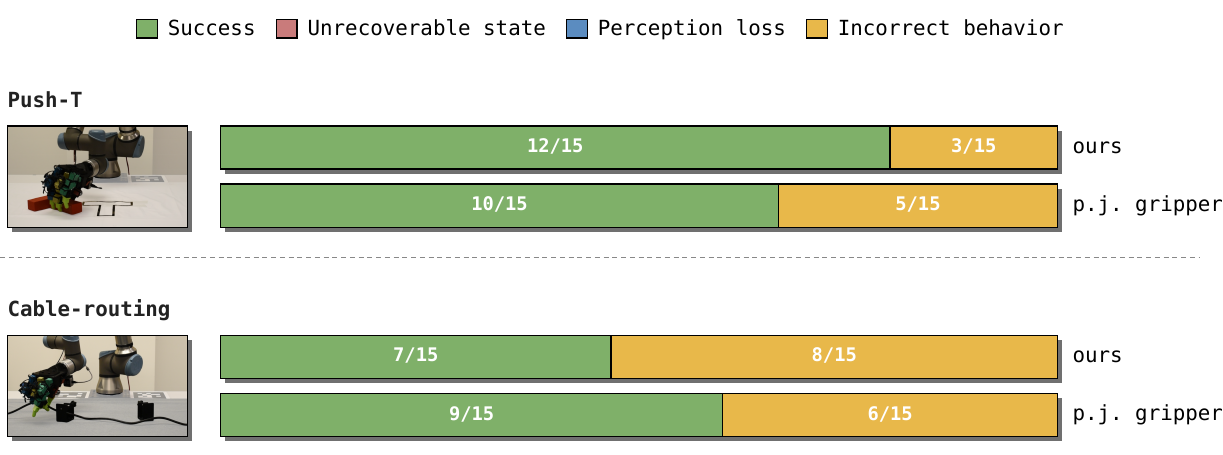}
\caption{\textbf{Failure-mode breakdown: self-collected tasks.} Per-trial outcomes for Push-T and Cable-routing. Each row compares execution with the dexterous hand policy and the parallel-jaw gripper policy.}
\label{fig:failures-self}
\end{figure}

We score each rollout as one of four outcomes; the legend colors match Fig.~\ref{fig:failures-web} and Fig.~\ref{fig:failures-self}.
\begin{itemize}
\item {\color[HTML]{7FB069}\textbf{Success}}: the rollout completes the task (per-task criteria are defined in \S\ref{sec:exp-capability} and \S\ref{sec:exp-embodiment}).
\item {\color[HTML]{C97B7B}\textbf{Unrecoverable state}}: the rollout enters a configuration from which the policy cannot make progress. Common cases include the manipulated object leaving the workspace (e.g., the apple rolling off the table in binning), or the robot getting stuck in a state where it cannot feasibly continue (e.g., the spoon is grasped at an orientation from which completing the rotations would push the wrist past its joint limit). Once this state is reached, no further closed-loop intent update can recover the rollout, and we stop the trial.
\item {\color[HTML]{5B8CC0}\textbf{Perception loss}}: the upstream perception pipeline loses the object during the rollout. Typical causes are the robot's hand or arm fully occluding the object for long enough that SAM~3.1's mask drifts off the object or DenseTrack3Dv2's tracked points reproject onto the wrong surface. Without a valid mask or 3D track, the intent model has no query to predict from and the policy stalls.
\item {\color[HTML]{E8B84A}\textbf{Incorrect behavior}}: the rollout neither succeeds nor enters a hard failure, but the policy executes a plan that does not actually complete the task. For example, in wiping, the marker remains visible on the whiteboard but the policy lifts the cloth and disengages anyway; or in binning, the predicted object flow points away from the target container, so the policy carries the object to the wrong location, often outside the camera bounds.
\end{itemize}

\subsection{Intent-Loss Scaling Extrapolation}
\label{app:intent-scaling-extrap}

To visualize extrapolation uncertainty, we fit fixed-exponent power laws $L(M)=c+aM^{-\alpha}$ to the five held-out intent-loss points from Fig.~\ref{fig:scaling}, where $M$ is the number of training clips. For each $\alpha\in\{0.05,0.10,0.15,0.20,0.25\}$, $c$ and $a$ are fit by least squares.

The candidates are nearly indistinguishable over the observed $1$K--$20$K clip range. The right panel extends the same fits $1000\times$ beyond the largest training set. Thus the current sweep supports the trend that more video helps but does not pin down a precise long-range scaling forecast.

\begin{figure*}[t]
\centering
\includegraphics[width=0.9\textwidth]{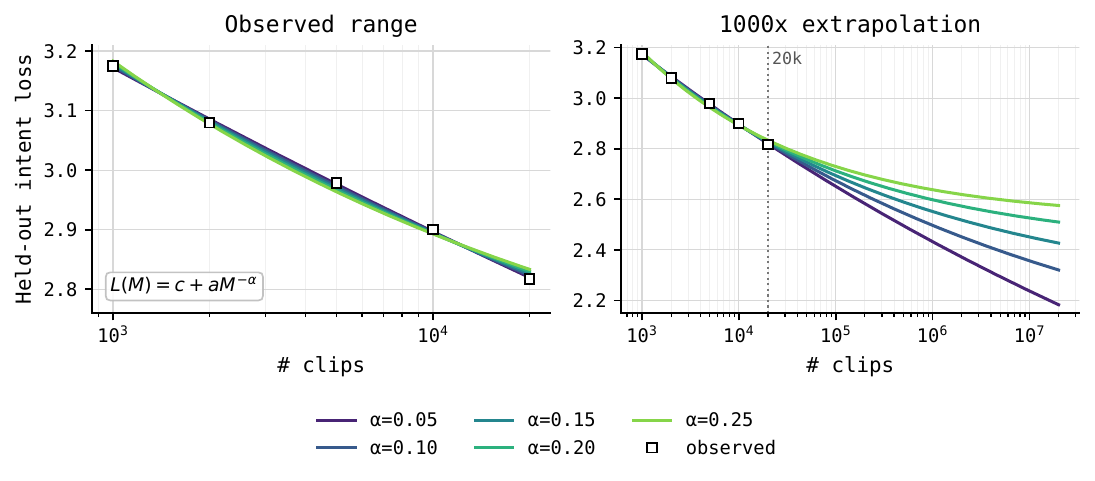}
\caption{\textbf{Power-law extrapolations of intent loss.}
Fixed-exponent fits to the held-out intent-loss points from Fig.~\ref{fig:scaling}. The measured range supports many similar fits, while extrapolation reveals substantial uncertainty in the long-range forecast.}
\label{fig:app-intent-scaling-extrap}
\end{figure*}

\subsection{Query-points ablation}
\label{app:query-points-ablation}

\begin{figure}[t]
\centering
\includegraphics[width=0.8\linewidth]{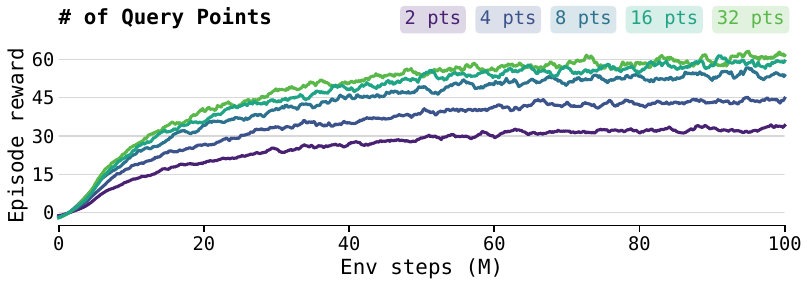}
\caption{\textbf{Query-points ablation.} Episode reward vs environment steps as we sweep the number of camera-visible query points $N$ the student policy receives.}
\label{fig:pc-ablation}
\end{figure}

We sweep the number of camera-visible query points $N$ the student policy receives during training (Fig.~\ref{fig:pc-ablation}). Episode reward continues to rise with more points but with diminishing returns. We pick $N = 16$ as our operating point: it captures most of the reward gain while keeping training fast for the intent model.

\end{document}